\documentclass{bmvc2k}

\usepackage{url}
\usepackage{graphicx}
\usepackage{amsfonts}
\usepackage{booktabs}
\usepackage{xcolor,colortbl}
\usepackage{wrapfig}
\usepackage{multirow}
\usepackage{multicol}

\usepackage{rotating} 
\newcommand{\emoji}[2][1.2em]{\raisebox{-0.2\height}{\includegraphics[height=#1]{#2}}}

\usepackage{stringenc}
\newcommand*{\uni}{}
\DeclareRobustCommand*{\uni}[1]{%
  \begingroup
    \StringEncodingConvert\x{%
      \pdfunescapehex{%
        00%
        \ifnum"#1<"100000 0\fi
        \ifnum"#1<"10000 0\fi
        \ifnum"#1<"1000 0\fi
        \ifnum"#1<"100 0\fi
        \ifnum"#1<"10 0\fi
        #1%
      }%
    }{utf32be}{utf8}%
    \everyeof{\noexpand}%
    \endlinechar=-1 %
  \edef\x{%
    \endgroup
    \scantokens\expandafter{%
      \expandafter\unexpanded\expandafter{\x}%
    }%
  }\x
}

\usepackage{xcolor}         
\usepackage{algorithm}
\usepackage{algpseudocode}

\usepackage{tcolorbox}                 
\definecolor{llmPrefix}{HTML}{253331}
\definecolor{llmQuestion}{HTML}{518E84}
\definecolor{llmOutput}{HTML}{253331}

\newenvironment{promptbox}[5][]        
{%
  \begin{tcolorbox}[left=1.5mm,right=1.5mm,top=1.5mm,bottom=1.5mm]
    \raggedright\small
    \ifx\relax#1\relax\else
      \begin{center}{\normalsize\textbf{#1}}\end{center}
    \fi
    {\color{llmQuestion}\textbf{Images:}}
    \fbox{\includegraphics[width=0.5\linewidth]{#2}}\\
    {\color{llmPrefix}\textbf{Instruction: } {\ttfamily #3}}\\[2pt]
    {\color{llmQuestion}\textbf{Questions: } {\ttfamily #4}}\\[6pt]
    {\color{llmOutput}\textbf{Answer: } {\ttfamily #5}}\\[6pt]
}{%
  \end{tcolorbox}
}

\definecolor{royal-blue}{RGB}{65, 105, 225}
\definecolor{vermilion}{RGB}{227, 66, 52}

\title{SynthFaces: Balanced Large Scale Human Dataset}

\addauthor{Sergey Karpukhin}{seriy.karp2@gmail.com}{1, 2}
\addauthor{Denis Bobkov}{dbobkov@hse.ru}{1, 2}
\addauthor{Kirill Morozov}{s02230196@gse.cs.msu.ru}{3}
\addauthor{Aibek Alanov}{aalanov@hse.ru}{1, 2}

\addinstitution{
 FusionBrain Lab, AXXX\\
 Moscow, Russia
}
\addinstitution{
 HSE University\\
 Moscow, Russia
}
\addinstitution{
 MSU\\
 Moscow, Russia
}

 \runninghead{Karpukhin \etal}{SynthFaces}

\def\etal{\emph{et al}\bmvaOneDot}

\begin{document}

\maketitle

\begin{abstract}

The pursuit of fair and equitable AI in human-centric tasks is critically hindered by the inherent biases and privacy risks of real-world datasets, which offer imbalanced representations of identity and often contain sensitive personal information. Consequently, state-of-the-art generative models trained on this skewed data exhibit unfair performance, particularly in identity-preserving generation. In this work we propose a pipeline for controllable generation of dataset with diverse faces for generative models training and finetuning. We achieve this by expanding foundational Arc2Face diffusion model with a flow matching generative model in the manifold of identity embeddings. We demonstrate that such approach achieves both goals of generative fidelity and label diversity not present in real human-centric datasets. As a result we collect and further contextualize a highly balanced identity-centric dataset of humans named SynthFaces. By design, our released images and identities are synthetic, reducing direct reliance on real personal images; we empirically check identity separation with a nearest-neighbor non-leakage metric. We purpose it for training and tuning of fair models for various downstream tasks, and showcase its usability in finetuning an existing adapter-based method for personalized generation.

\end{abstract}

\section{Introduction}
\label{sec:intro}

Human image generation is one of the most common applications of modern generative models. The field originated with unconditional generators such as StyleGAN \cite{fu2022stylegan}, which allowed to sample novel faces without any control, and was later expanded to text-to-image diffusion models and adapter-based approaches (e.g., PhotoMaker, SDXL-based personalization), which allow conditioning on a novel face from a single image \cite{rombach2022high, ye2023ip, li2024photomaker, wang2024instantid} and combined with prompt allowed to place human in different context. Unfortunately, most released face-conditional models report averaged quality, which conceals an important aspect of such methods - \textit{do they perform equally for different human subgroups}? In this work we show that this is not the case, and try to address this problem with our curated dataset \textbf{SynthFaces} collected with focus on balanced human representation.

SynthFaces dataset collection pipeline is designed to address a set of three core problems. \textbf{1) Label bias of existing datasets:} both existing datasets with human faces, such as FFHQ, SHHQ \cite{fu2022stylegan}, EasyPortrait \cite{kvanchiani2023easyportrait} and CosmicMan-HQ \cite{li2024cosmicman} and generative samplers exhibit severe category imbalance - both unconditional samplers such as StyleGAN \cite{melzi2023gandiffface}, as well as by conditional samplers e.g., Arc2Face \cite{papantoniou2024arc2face}, PhotoMaker \cite{li2024photomaker} produce or reflect skewed distributions over demographic attributes (age, race, gender). \textbf{2) Face-similarity quality of conditional samplers:} when evaluated per demographic group, conditional personalization models show uneven face preservation quality, which is natural given first problem; \textbf{3) Usage of real people identities:} existing real datasets, especially large-scale ones, may include identities that did not give consent for deep learning model training or may include identities from under-age groups, using which may be completely unethical and certain datasets have been revoked from public access for that reason \cite{kemelmacher2016megaface, guo2016ms}. 

Additionally, we aim to fulfill properties that are largely required for modern human-centric datasets - such properties are mentioned in Table \ref{tab: high-level data}, specifically we aim to provide \textbf{Captions} for images, describing both humans and surrounding context, collected images are \textbf{ID-grouped}, meaning that there are multiple images with same person, and lastly we aim to obtain \textbf{HQ} image with high resolution. These properties may be missing for certain types of datasets mentioned in table. For example  under \textbf{Raw Images} we denote large-scale raw collections obtained via web-scrapping, are large enough to include all types of faces, however they might still require label-balancing strategy, as well image filtering to obtain high-quality images. \textbf{HQ Captioned} datasets are high quality, but they balance in human representation as we show later, as well as typically do not include multiple images of same person. \textbf{Face Recognition} datasets are typically also without captions, low resolution and also skewed in terms of human representation. In comparison our collected dataset satisfies all of these properties, while also using released images and identities that are synthetic and empirically screened for identity separation.

\begin{table}[htb]
    \caption{High level comparison of dataset types, in columns \textbf{Caption} denotes that images are paired with prompts, \textbf{Balanced labels} that dataset is balanced over selected demographic labels, \textbf{ID-grouped} that multiple images per person are provided and \textbf{HQ} suggests that images are high resolution and visually high quality}
    \centering
    \resizebox{\textwidth}{!}{
    \begin{tabular}{l c c c c c c }
        \toprule
        \textbf{Type} & \textbf{Dataset} &\textbf{ Captions} & \textbf{Balanced labels} & \textbf{ID-grouped} & \textbf{Private} & \textbf{HQ} \\
        \midrule
        Raw Images & VGG-Face, WebFace & \emoji{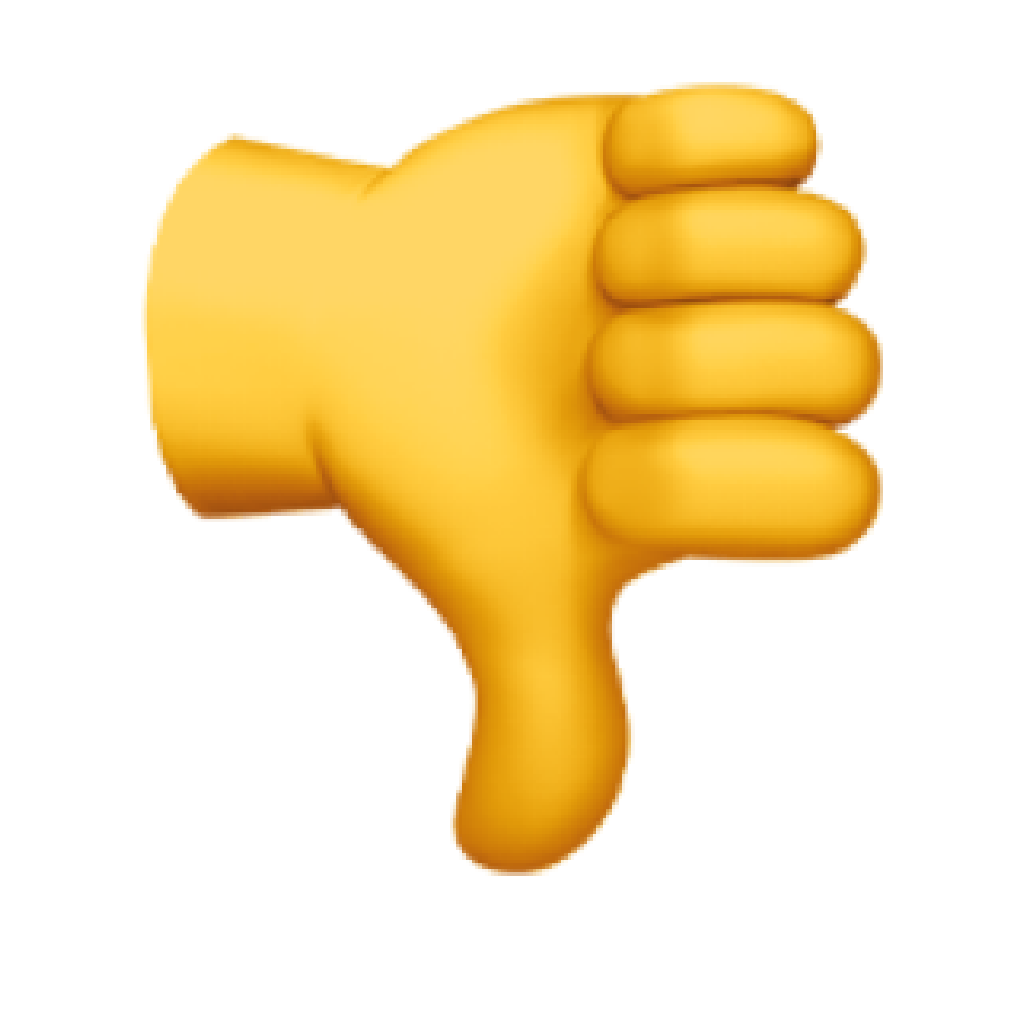} no & \emoji{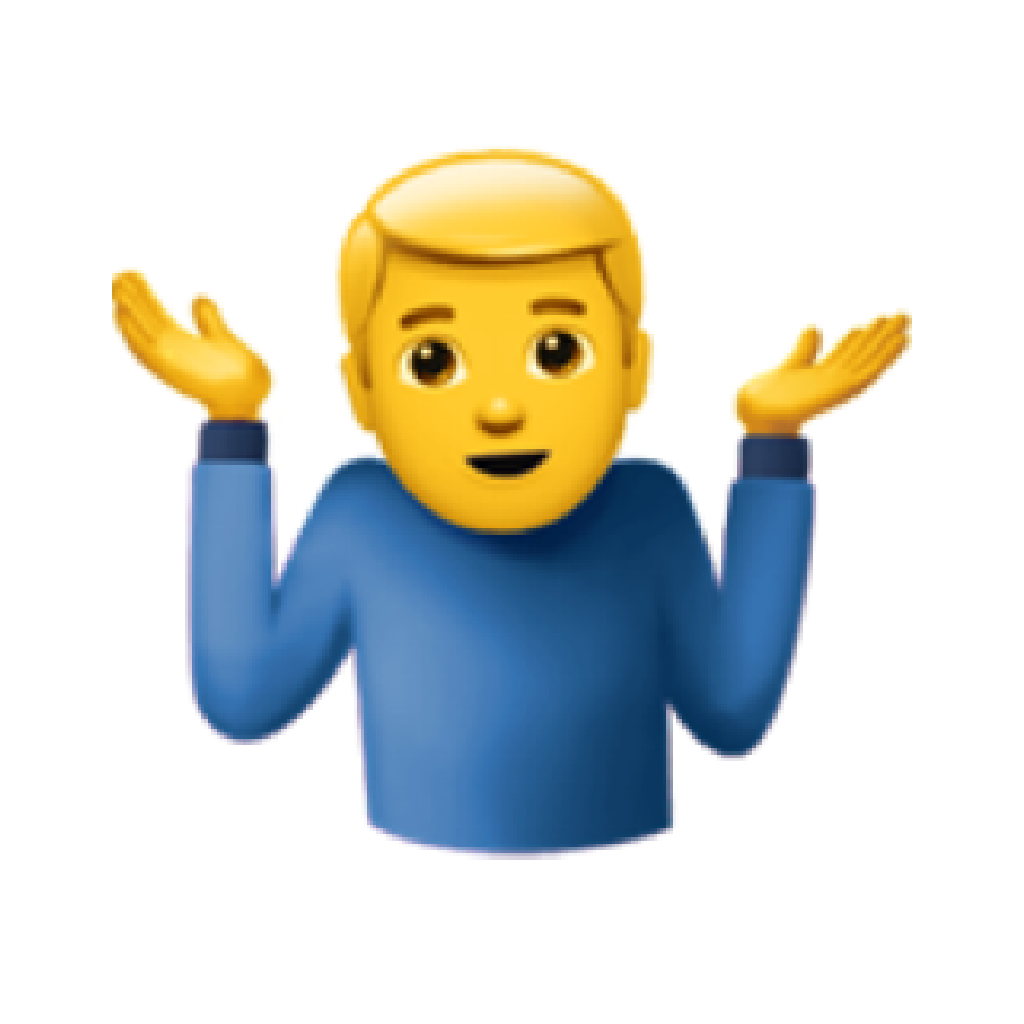} partially & \emoji{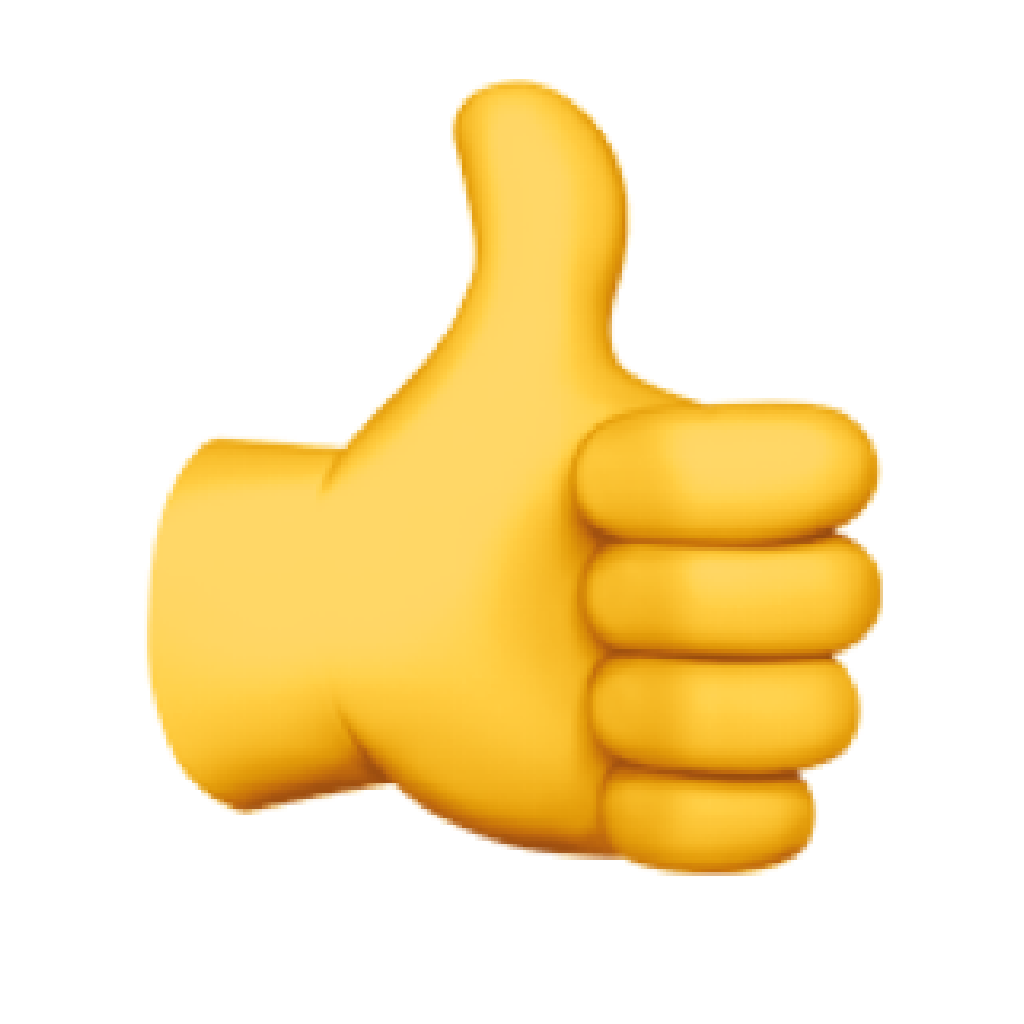} yes &  \emoji{figures/emojis/thumbs-down.png} no & \emoji{figures/emojis/man-shrugging.png} partially \\
        HQ Captioned & Cosmic, DeepFashion & \emoji{figures/emojis/thumbs-up.png} yes  & \emoji{figures/emojis/thumbs-down.png} no & \emoji{figures/emojis/thumbs-down.png} no & \emoji{figures/emojis/thumbs-down.png} no & \emoji{figures/emojis/thumbs-up.png} yes  \\
        Face Recognition & DCFace, HyperFace & \emoji{figures/emojis/thumbs-down.png} no & \emoji{figures/emojis/thumbs-down.png} no & \emoji{figures/emojis/thumbs-up.png} yes & \emoji{figures/emojis/man-shrugging.png} partially & \emoji{figures/emojis/thumbs-down.png} no  \\
        \midrule
         & \textbf{SynthFaces (ours)} & \emoji{figures/emojis/thumbs-up.png} yes & \emoji{figures/emojis/thumbs-up.png} yes  & \emoji{figures/emojis/thumbs-up.png} yes  & \emoji{figures/emojis/thumbs-up.png} yes  & \emoji{figures/emojis/thumbs-up.png} yes \\
        \bottomrule
    \end{tabular}
    }
    \label{tab: high-level data}
\end{table}

\textbf{SynthFaces} addresses these problems by introducing controllable label diversity into identity representation sampling and by releasing synthetic identities whose separation from real training identities is checked through ID-NonLeak. The resulting dataset is balanced across selected demographic labels, ID-grouped, captioned, and contextualized for high-resolution human generation. We systematically study label balance, non-leakage, image quality, and downstream usefulness through generative fine-tuning, face recognition fairness, and demographic VQA in Section~\ref{sec: downstream_application}.

\section{Related Works}

\paragraph{Synthetic face datasets} have been studied primarily for face recognition \cite{kim2023dcface, bae2023digiface, li2024id}. Existing pipelines combine GAN samples with per-identity adaptation \cite{kvanchiani2023easyportrait,ruiz2023dreambooth}, or exploit the conditional space of face foundation models such as Arc2Face \cite{papantoniou2024arc2face}. HyperFace \cite{shahreza2024hyperface} explores this hyperspherical space directly, but its generated identities remain demographically imbalanced. Arc2Face and Vec2Face~\cite{wu2025vec2face} instead perturb real ArcFace embeddings with PCA, requiring real identity embeddings at inference time and offering no explicit separation from them. ID-Flow learns a controllable prior over the identity manifold, samples without reference embeddings at inference time, and evaluates separation using ID-NonLeak (Table~\ref{tab:fid-baselines}).

\paragraph{Model fairness} seeks to reduce systematic performance differences across demographic subgroups \cite{mehrabi2021survey, li2023survey}. Our approach intervenes at data construction time by controlling the demographic distribution of sampled identities.

\begin{figure}[bth!]
    \centering
    \includegraphics[width=0.95\textwidth]{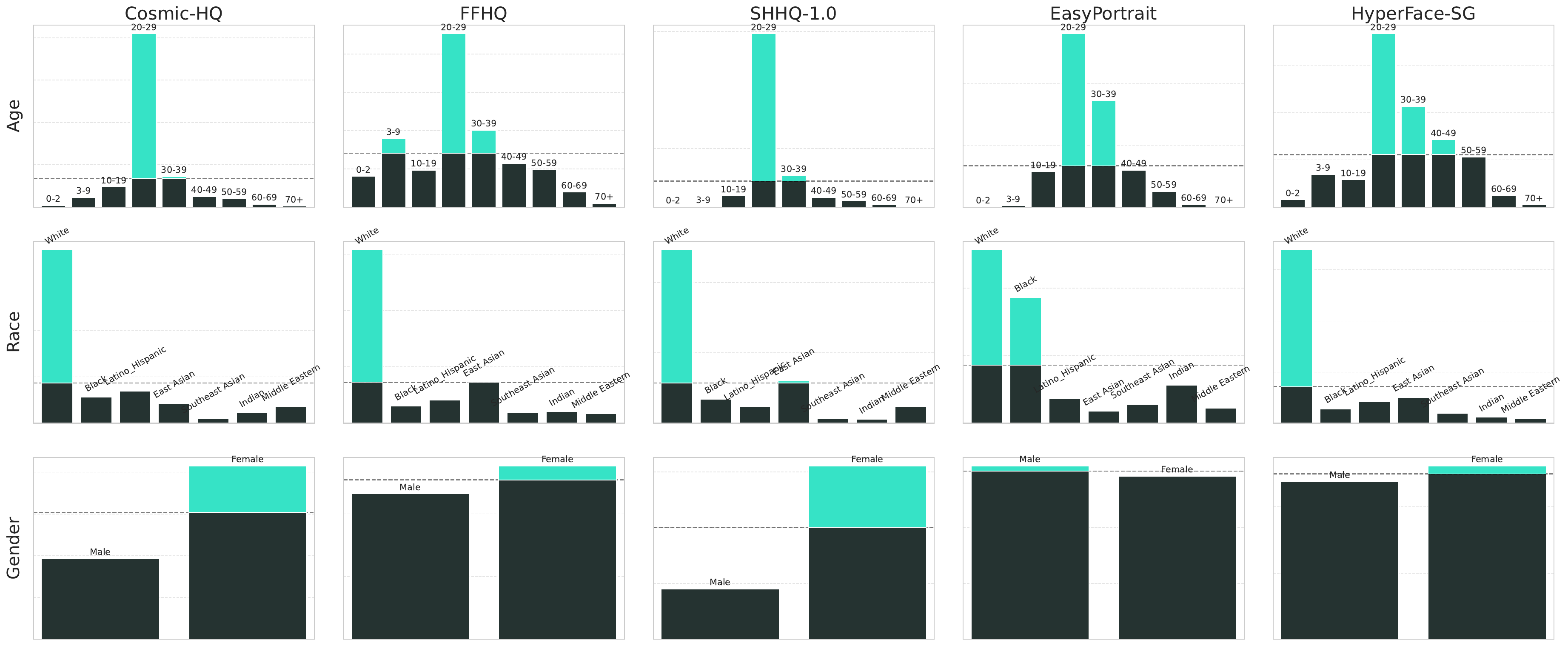}
    \caption{Distribution of labels across different categories in some of conventional datasets with humans. Dashed lines denotes mean value across labels, color highlights values above mean}
    \label{fig:intro-fairface-datasets}
\end{figure}

\paragraph{Identity-preserving generation}
Identity-preserving generation conditions image synthesis on a reference face. Introduced with adapter-based methods - IP-Adapter-FaceID~\cite{ye2023ip}, PhotoMaker~\cite{li2024photomaker}, PuLID~\cite{guo2024pulid}, and InstantID~\cite{wang2024instantid}; InfiniteYou~\cite{jiang2025infiniteyou} further improves fidelity and flexibility. Recent in-context models such as Qwen-Image~\cite{wu2025qwen} and Flux-2~\cite{flux-2-2025} can preserve identity across varied contexts, whereas per-subject methods such as DreamBooth~\cite{ruiz2023dreambooth} require fine-tuning for every new identity. We use these models for contextualization after independently controlling the identity distribution.

\section{Motivation}

\subsection{Existing datasets biases}

To quantify label bias in existing datasets, we annotate them with FairFace  classifier \cite{karkkainen2021fairface} and in Figure \ref{fig:intro-fairface-datasets} show that all of the presented datasets demonstrate heavy bias towards certain labels. Training on such data without empirical balancing strategies can be harmful for generative model fairness and diversity.

\subsection{Evaluating unfairness in human-centric models}

To expose subgroup differences hidden by aggregate metrics, we construct \textbf{FairBench}, a probe set sampled uniformly over FairFace label combinations. Figure~\ref{fig:facesim-unfairness} shows substantial demographic variation in identity preservation, consistent with the dataset skew in Figure~\ref{fig:intro-fairface-datasets}. Section~\ref{sec: downstream_application} evaluates whether balanced training data also improves subgroup parity in face recognition and demographic VQA.




\section{Dataset construction}

\begin{figure}[!bth]
    \centering
    \includegraphics[width=0.85\textwidth]{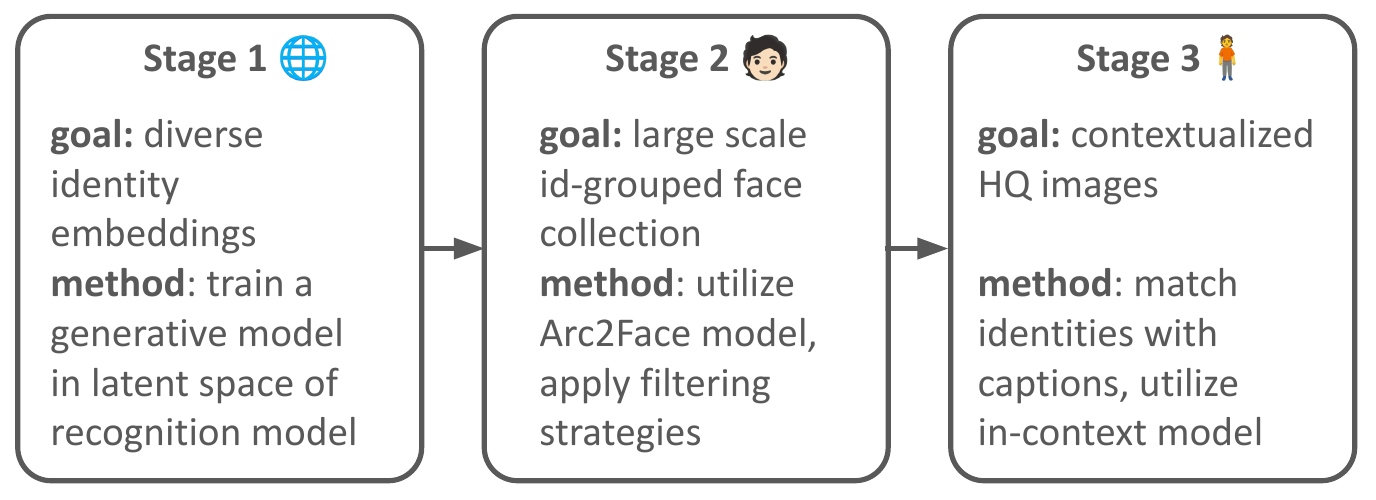}
    \caption{Dataset multi-stage collection pipeline.}
    \label{fig:stages}
\end{figure}

In order to build a dataset of synthetic released identities and images, we follow a 3 stage design described in Figure \ref{fig:stages}. Each stage is motivated by a specific problem of existing pipelines - at stage 1 we ensure that dataset is private while also free of skewed label distribution, at stage 2 we construct maximum fidelity faces from diverse identity representation, and at stage 3 we achieve high resolution and context coupled with captions.

\subsection{Learning to sample identities}

Identity embeddings lie on a unit hypersphere, so treating them as unconstrained Euclidean vectors ignores their geometry. We instead learn a geodesic flow-matching model directly on this manifold.

\begin{figure}[t]
    \centering
\includegraphics[width=0.95\linewidth]{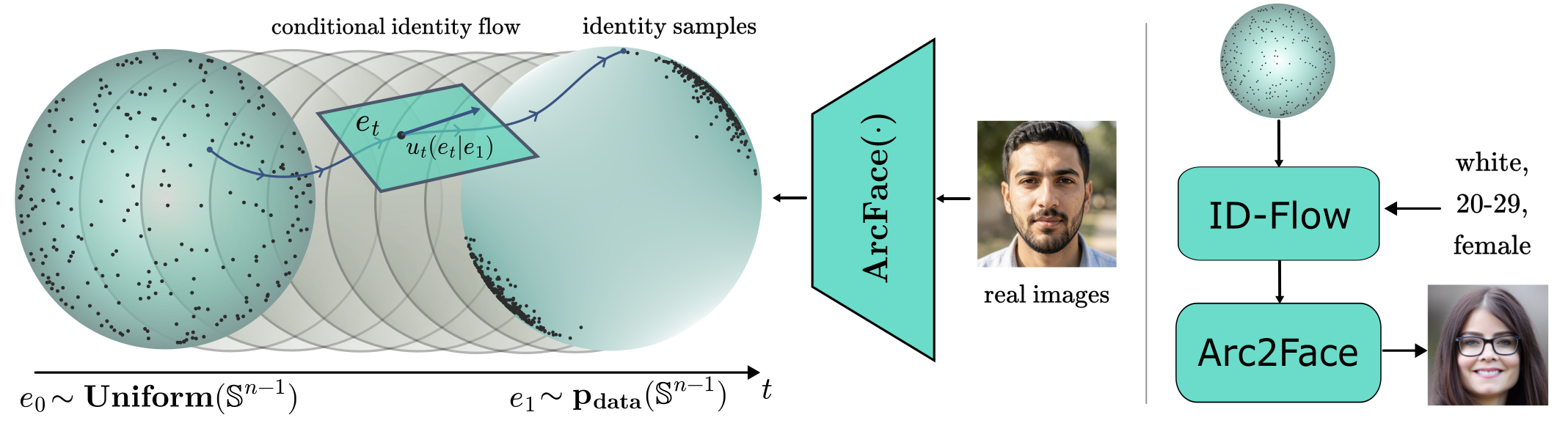}
    \caption{Left - Conditional identity flow between uniform points on sphere and ArcFace embeddings used for training, constrained to tangent hyperplanes, right - inference scheme together with Arc2Face at stage 2, trained ID-flow samples identity embedding with given control signals and Arc2Face samples image}
    \label{fig:id-embedding-diff}
\end{figure}

We aim to train a generative model in latent space of identity embeddings. Natural observation is that these embeddings are all unit length and therefore exists on a $n$-dimensional sphere. Some prior works tried to explore this idea, for example in \cite{shahreza2024hyperface} authors pose problem of sampling diverse embeddings as optimization problem on sphere, and in \cite{dosi2025harmonizing} authors construct von Mises–Fisher based diffusion model that respects manifold geometry. In this work we make \textit{a novel adaptation of non-Euclidean flow matching} to construct a generative model on identity latents. Given $\{e_{i}\}^N_{i=1} \in \mathbb{S}^n$, where $\|e_i\|=1$ for $\forall i$, we aim to train a $v_\theta(\cdot)$ to approximate velocity field of the tangent bundle $T\mathcal{M}$ of sphere manifold between two probability distributions -  prior uniform on sphere $\mathbf{Uniform}(\mathbb{S}^{n-1})$ and distribution of identity embeddings $p_{1}(\mathbb{S}^{n-1})$. We follow standard definitions and define geodesic conditional flow as $\phi(x_t|x_1) = \exp(k(t) \log_{x_0}(x_1))$, where $t \in [0, 1]$. Functions $\exp_{x_0}(\cdot)$ is a projection from sphere to tangent space $M_{x_0}$ and $\log_{x_0}(\cdot)$ is the inverse. These projection are explicitly defined for known manifolds such as sphere, which allows to obtain training target without requiring simulation \cite{lipman2024flow}. Overall scheme of flow construction and resulting pipeline with trained model is presented in Figure \ref{fig:id-embedding-diff}, and the training objective is given in Eq.\ref{eq:gcfm}. For definitions of projections and background on flow matching check Appendix \ref{sec: sphere projections}

\begin{equation}
   L_{GCFM}(\theta) = \mathbb{E}_{t, e_0, e_1}\|v_\theta(e_t, c) - \frac{d}{dt}e_{t|1}\|_2^2 
   \label{eq:gcfm}
\end{equation}

Since we aim to train a controllable conditional model, along with interpolated $e_t$ learned velocity will accept as input a set of labels $c = [c_1, c_2, c_3]$, representing face. These labels are obtained through training data annotation and represent general face features, such as race, age and gender. During inference we use these to construct diverse dataset by applying classifier-free guidance \cite{ho2022classifier}. We stick with simple multilayer MLP architecture with skip-connections and apply guidance scale 2.0 during sampling of identities. ID-Flow is trained on approximately 50k deduplicated ArcFace embeddings extracted from CosmicMan-HQ, FFHQ, and EasyPortrait; full data collection and model details are given in Appendix~\ref{sec:model_details_and_ablations}. 

\begin{table}[!htb]
    \centering
    \caption{Baseline comparison of Stages 1 and 2. Lower FID is better and higher Precision/Recall/F1/ID-NL scores are better. ID-NL (ID-NonLeak) measures average nearest-neighbor separation from real training identities.}
    \label{tab:fid-baselines}
    \resizebox{\textwidth}{!}{
    \begin{tabular}{lccccccc}
        \toprule
        Category & Method & FID-i $\downarrow$ & FID-e $\downarrow$ & Prec. $\uparrow$ & Rec. $\uparrow$ & F1 $\uparrow$ & ID-NL $\uparrow$ \\
        \midrule
        \multirow{2}{*}{Naive} 
            & uniform         & 96.53  & 0.277  & 0.011 & 0.95 & 0.02  & --    \\
            & *interpolation  & 30.59  & 0.035  & 0.765 & 0.622 & 0.68 & 0.43  \\
            & *PCA  & 51.18  & 0.048  & 0.29 & 0.88 & 0.44 & \textbf{0.76}  \\
        \midrule
        \multirow{4}{*}{Model}
            & StyleGAN        & 45.8 & \underline{0.111}  & \underline{0.711} & 0.173  & 0.28  & --    \\
            & Arc2Face        & \textbf{37.3} & 0.118  & 0.701 & \underline{0.185} & \underline{0.29}  & --    \\
            & SDXL-Photomaker & 73.9 & 0.207  & \textbf{0.73} & 0.01 & 0.02  & --    \\
            & HyperFace       & 61.1 & 0.13  & 0.57 & 0.17 & 0.26 & --    \\ 
        \midrule
         \textbf{Ours}   & \textbf{ID-Flow} & \underline{39.9} & \textbf{0.073} & 0.35 & \textbf{0.75} & \textbf{0.48} & \underline{0.741}\\ 
        \bottomrule
    \end{tabular}
    }
    \vspace{0.5em}

    {\small *Interpolation and PCA baselines use real training embeddings.}
\end{table}

We compare against other baselines in terms of a) face image quality b) label diversity. In Table \ref{tab:fid-baselines} these baselines are: parameter-free or \textbf{naive} such as uniform sampling on sphere, interpolation of embeddings and PCA based sampling, following procedure described in \cite{wu2025vec2face}, unconditional StyleGAN and conditional \textbf{model}-based samplers - Arc2Face and Photomaker that were used with synthetic identities, as well as recent HyperFace method for synthetic generation. We report image FID (FID-i) with face hold-out set, and generative metrics for identity latents (FID in embedding space (FID-e), Precision, Recall, F1) \cite{kynkaanniemi2019improved}. During sampling with ID-flow we do not utilize labels from test set, but sample labels from train data statistics for fair comparison. ID-Flow gives the best FID-e, recall, F1, and ID-NonLeak among learned samplers, while keeping image FID close to Arc2Face.

\begin{figure}[bth!]
\includegraphics[width=0.95\linewidth]{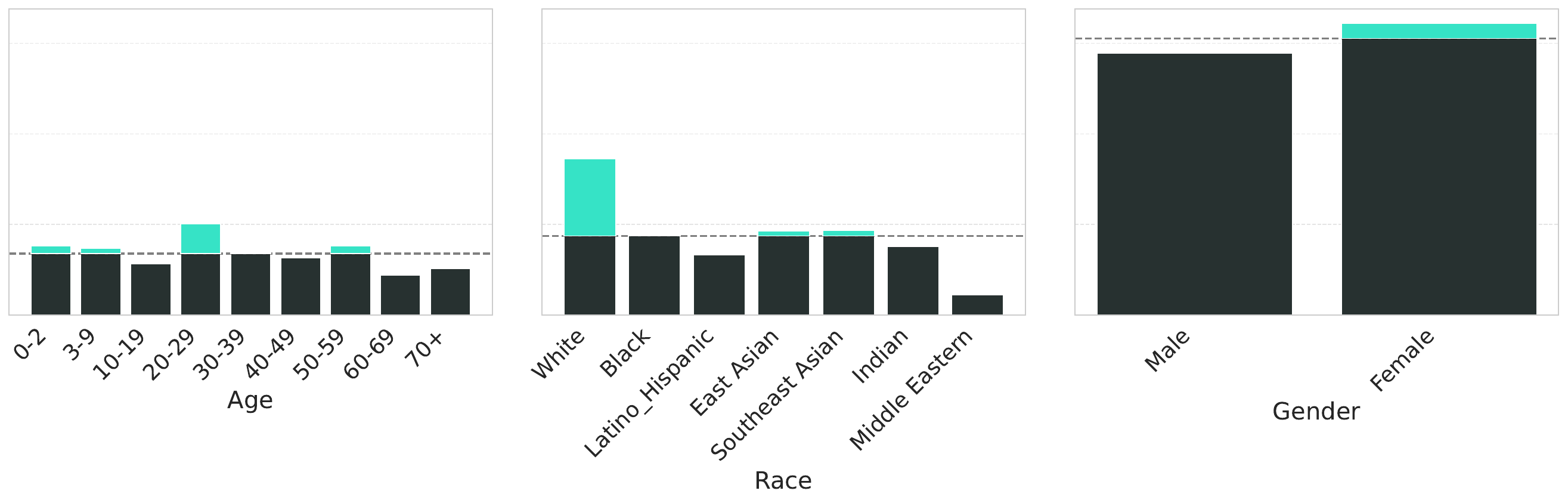}
    \caption{Distribution of different label categories data generation by ID-Flow. Compared to conventional datasets, our method offers significant data balance for all major characteristics of persons}
    \label{fig:model_samples_hist}
\end{figure}

Compared to other models, our method offers superior identity diversity, since other baseline samplers inherit mode collapse property of synthetic data. Despite strong results of interpolation,  we show that it results in leakage of training data into samples (ID-NonLeak) - we measure identity leakage over the training set as an average cosine distance to nearest neighbor from training data - samples from model are almost orthogonal to closest samples from train data, while interpolated samples remain in close angular proximity.

\begin{wraptable}{r}{0.35\linewidth}
    \caption{Label diversity}
    \label{tab: diversity}
    \begin{tabular}{lc}
        \toprule
        Method & $KL$ \\
        \midrule
            FFHQ & 0.449 \\
            SHHQ & 0.533 \\
            EasyPortait & 0.508 \\
            CosmicHQ-1.0 & 0.576 \\
            HyperFace-10k & 0.49 \\
            PCA & 0.75 \\
            SynthFaces (ours) & \textbf{0.048} \\
        \bottomrule
    \end{tabular}
\end{wraptable}

\begin{figure}[bth!]
    \centering
\includegraphics[scale=0.53]{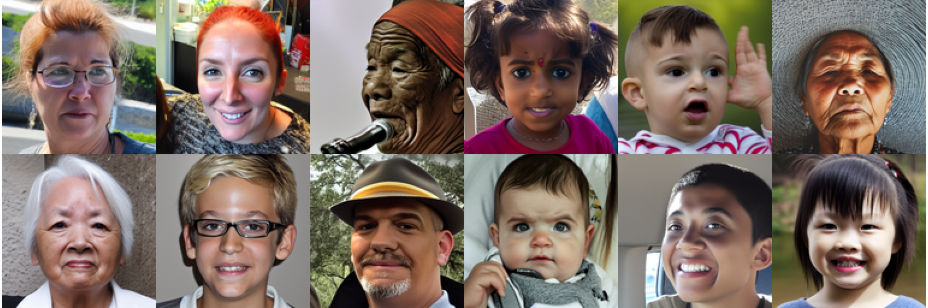}
    \caption{Diverse samples from ID-Flow. Dataset was sampled with uniform label controls, ensuring greatest variance of identity representations}
    \label{fig:model_samples}
\end{figure}

In Table \ref{tab: diversity} we numerically compare diversity of generated data when control labels provided to the model across different categories are selected uniformly against baselines reported in Fig. \ref{fig:intro-fairface-datasets}. We evaluate diversity through KL-divergence with uniform distribution $KL(p_{model}(y)||U(y))$ averaged across label categories. Additionally we show visual histograms of generated data samples with our method in Figure \ref{fig:model_samples_hist}, where it can be observed that compared to baselines in Fig. \ref{fig:intro-fairface-datasets} this stage pipeline of ID-flow with Arc2Face allows to achieve highly balanced data, while also surpassing baselines in terms of face quality. Samples of identities obtainable at this stage are provided in Fig. \ref{fig:model_samples}


After training an ID-flow, we start by sampling a large amount of unique identity representations, resulting in 28 thousands embeddings. These embeddings as noted by \textbf{Stage 2} plan, are turned into face images, and various filtering strategies are applied in order to enhance output of this stage and get rid of potential artifacts present in images.

\paragraph{Image statistics filtering} we start by analyzing distribution of images characteristics generated by Arc2Face. We evaluate conventional computer vision characteristics of images, such as \textbf{entropy} and \textbf{saturation}. We observe that tail of entropy distribution corresponds to heavily stylized images, low saturation values correspond to black and white images, and high saturation mostly coincides with faces that have unnatural amount of color/light or make up. Therefore we drop around p\% of images from tails, removing approximately 3k unsatisfactory images. Qualitative examples are provided in supplementary Figure~\ref{fig:stats-filter}.

\begin{figure}[!thb]
    \centering
\includegraphics[width=0.95\linewidth]{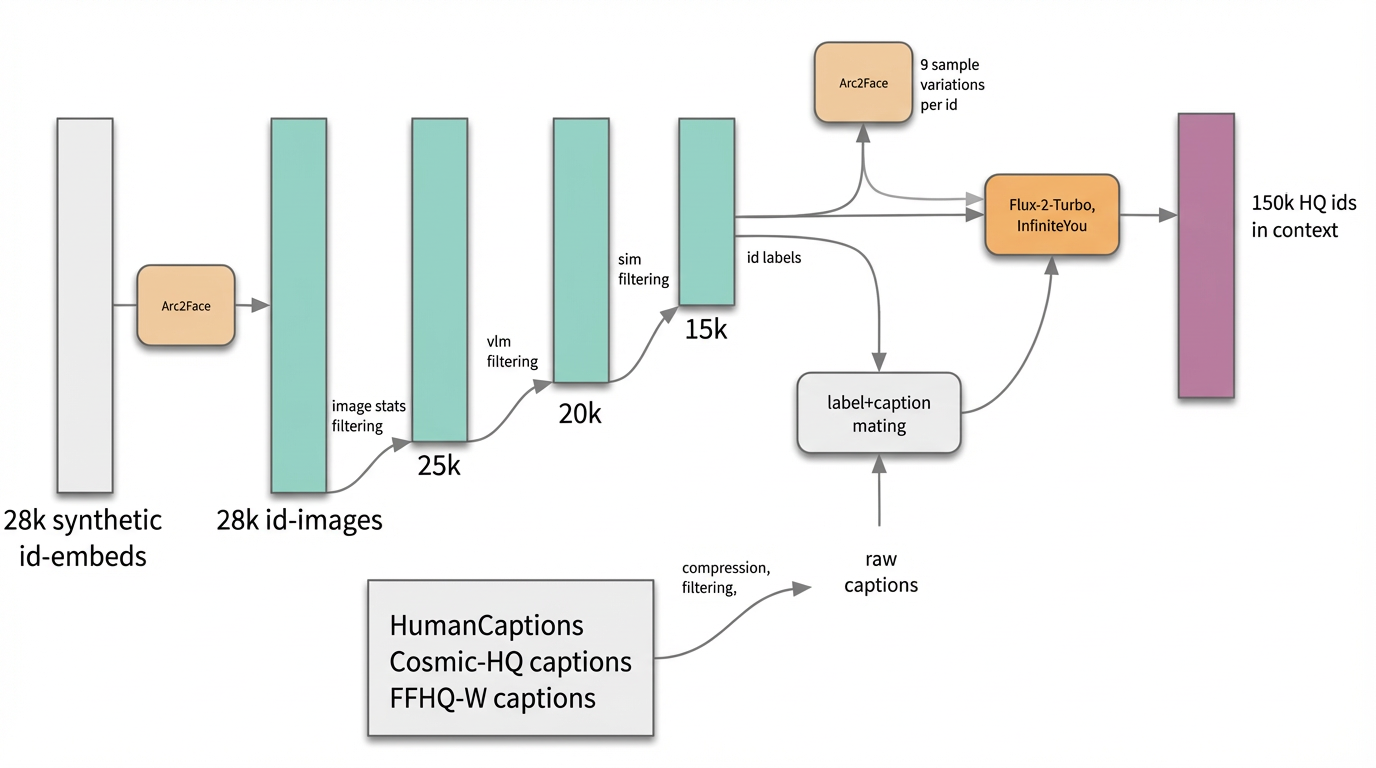}
    \caption{Data filtering pipeline from Stage 2 to contextualization - in order to minimize amount of corrupted images with faces we 1) remove images in long tails of statistics 2) detect images with artifacts via vision-language model 3) threshold pairwise similarity, ensuring variability of identities in all categories, resulting in the end with approximately 150,000 final images}
    \label{fig:pipeline}
\end{figure}
\subsection{Data curation pipeline}

\paragraph{VLM filtering} After basic image statistics filtering we utilize multiple Qwen-VL models to assess obtained images from two perspectives 1) does the person's face align with labels that were used to generate them 2) does the person's face contain any visual artifacts. Based of VLM answers we drop all misaligned images and images that contained faces with visual defects, keeping samples which by majority voting of VLMs are assessed as well-aligned. To make sure that only relevant artifacts are dropped, we prompted VLM to reason about noticed artifacts in the images and dropped images only if the reasoning contained specific key-words. Detailed example of VLM prompting is provided in Appendix \ref{sec: vlm_filter_stage}. 

\begin{table}[!htb]
    \centering
    \caption{Comparison of metrics before and after filtering Stage 2 images. F1/Precision/Recall/FID and KL divergence same as in previous tables, KL is split for separate categories analysis. Note that metrics are computed with skewed test set of face images}
    \label{tab:before_after_metrics}
    \resizebox{\textwidth}{!}{
    \begin{tabular}{lcccccccc}
        \toprule
        & \textbf{F1} & \textbf{Precision} & \textbf{Recall} & \textbf{FID$_e$} & \textbf{FID$_i$} & \textbf{KL-age} & \textbf{KL-race} & \textbf{KL-gender} \\
        \midrule
        \textbf{Before} & 0.496 & 0.425 & 0.596 & 0.163 & 65.59 & 0.000108 & 0.000042 & 0.000001 \\
        \textbf{After} & 0.444 & 0.359 & 0.580 & 0.127 & 57.66 & 0.121146 & 0.006530 & 0.003619 \\
        \bottomrule
    \end{tabular}
    }
\end{table}

\paragraph{Similarity filtering} Additionally, we made sure that resulting identities are not only diverse in terms of labels, but also in terms of embeddings themselves. For that purpose we evaluated distribution of nearest neighbors in the resulting identities and observed a long tail which is not present in typical datasets. After per-label analysis these higher similarity tails occurred for specific label combinations that were particularly rare in training data used for ID-flow training - e.g. rare race groups or age groups. Therefore we filter by similarity threshold for these specific groups, while also keeping a proportion of data across labels close to uniform. After this stage with obtain highly diverse identities, superior to diversity in other baselines - 15 thousand filtered identities and respective labeled images remain. Detailed analysis of this stage results is given in Appendix \ref{sec: sim_filter_details}. The overall image statistics before and after all filtering strategies applied at this stage are given in Table \ref{tab:before_after_metrics}. Filtering improves image quality metrics (FID$_e$ and FID$_i$) and removes visible artifacts, but it also reduces F1/precision/recall and worsens KL-age because the most problematic tails are not perfectly uniform distributed across demographic labels.

\begin{table}[!htb]
    \centering
    \caption{Text-conditioned Flux.2 sampling versus ID-Flow ($N{=}1000$ each). Label KL is the mean KL divergence from a uniform distribution over age, race, and gender; ID-div. is mean nearest-neighbor cosine distance.}
    \label{tab:flux-direct-sampling}
    \begin{tabular}{lcc}
        \toprule
        \textbf{Method} & \textbf{Label KL $\downarrow$} & \textbf{ID-div. $\uparrow$} \\
        \midrule
        Flux.2 T2I, unconditional & 0.891 & 0.257 \\
        Flux.2 T2I, balanced prompts & 0.101 & 0.201 \\
        \textbf{ID-Flow (ours)} & \textbf{0.037} & \textbf{0.751} \\
        \bottomrule
    \end{tabular}
\end{table}

After filtering, we additionally test whether balanced prompting of a state-of-the-art text-to-image model can replace explicit identity-manifold sampling. We generate 1,000 Flux.2-dev-Turbo seed faces using either a generic prompt or cycled demographic descriptors and compare them with 1,000 ID-Flow samples. Table~\ref{tab:flux-direct-sampling} shows that balanced prompting improves Flux.2 label balance, but remains 2.7$\times$ farther from uniform than ID-Flow (0.101 vs.\ 0.037) and yields 3.7$\times$ lower identity diversity (0.201 vs.\ 0.751). Prompt-level demographic control therefore does not provide the balanced and diverse identity distribution required for dataset construction.

\subsection{Obtaining HQ images}

\paragraph{Caption and label matching} 

\begin{wraptable}{r}{0.55\linewidth}
    \caption{Comparison of HQ dataset metrics with SynthFaces - our dataset offers superior quality over Cosmic-HQ, while allowing richer contextualization without overly large faces}
    \label{tab: stage3-metrics}
    \begin{tabular}{lccc}
        \toprule
        & \textbf{FaceSize} & \textbf{IR} & \textbf{CLIP} \\
        \midrule
        Cosmic-HQ & 0.22 & -0.69 & 31.20 \\
        FluxSynID \cite{ismayilov2025fluxsynid} & 0.61 & 1.54 & 34.07 \\
        \textbf{SynthFaces} & 0.29 & 1.38 & 33.04 \\
        \bottomrule
    \end{tabular}
\end{wraptable}

In order to contextualize images, we collected captions of images from real world datasets like Cosmic-HQ and FFHQ-in-the-Wild and collections of human related texts like HumanCaption \cite{dai2025humanvlm}. We filtered prompts to exclude any details related to attributes that were controlled during identity sampling, including details with description of facial features. After that we roughly annotated these captions into subgroups based of their context, main factor being suitable age, e.g. prompts for child, prompts for adults, etc. After that we randomly sampled captions for each image given its label, matching label with classes with captions that are suitable for these face label values by looking for specific keywords that were used as person descriptors. Details of algorithm are given in Appendix \ref{sec: label_matching_alg}. 

\begin{figure}[!htb]
    \centering
\includegraphics[width=0.91\linewidth]{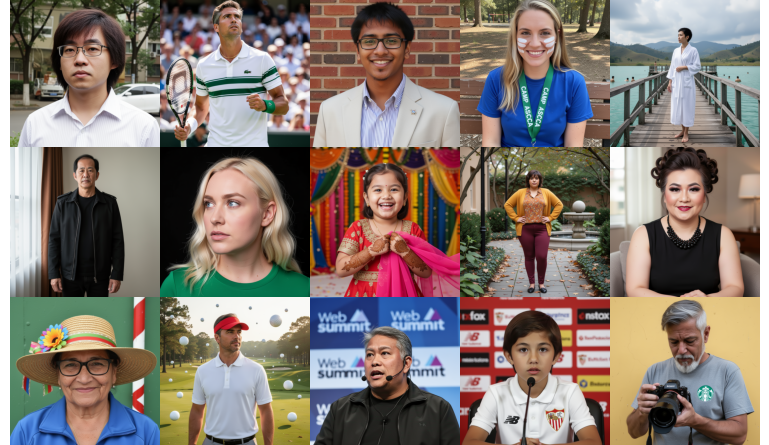}
    \caption{SynthFaces dataset sample examples - after filtration and caption matching identities are contextualized with Flux-based methods, resulting in total 150,000 high-quality contextualized images with approximately 10 images per identity }
    \label{fig:dataset-canvas}
\end{figure}

Examples of ID-grouped images - multiple contextualized images per synthetic identity, as well as images with more variance in terms of face position in image, are shown in Appendix~\ref{sec: synthface samples}.

\paragraph{Identity contextualization} In order to unite curated identities with captions, we utilize state of the art personalization models - InfiniteYou and in-context model Flux 2, first one aims for consistency while second one allows more fidelity and variability. In order to enhance realism of images generated with InfiniteYou we attach additional LoRAs for realism and deblurring, and to speed up generation with Flux 2 we utilize Turbo checkpoint that from experiments turned to be the best amongst existing models for personalized generation of humans, see qualitative comparison in Appendix \ref{sec: incontext_ablation}. As a result, we obtain a large scale, high quality diverse dataset of synthetic released identities, consisting in total of 150,000 captioned high-resolution images of varying aspect ratios, with approximately 10 images per identity. Resulting samples from this dataset are given in Fig. \ref{fig:dataset-canvas}. 

Additionally, we evaluate resulting SynthFaces images in terms of image quality by \textbf{ImageReward(IR)} metric \cite{xu2023imagereward}, text alignment by \textbf{CLIP} and ratio of size of face detection box to size of the image side. Compared with the baselines in Table~\ref{tab: stage3-metrics}, SynthFaces improves over Cosmic-HQ on ImageReward and CLIP while keeping faces smaller than FluxSynID, whose document-formatted images often lack background context.

\section{Downstream applications}
\label{sec: downstream_application}

We evaluate whether SynthFaces data yields fairness benefits in distinct downstream tasks: finetuning personalized generation pipeline, face recognition (FR) verification and demographic attribute classification. Fine-tuning tests usefulness for id-preserving generation, FR tests whether balanced synthetic identities reduce verification disparity, and VQA tests whether balanced human images help demographic attribute recognition.

\subsection{ID-preserving generation}
\label{subsec: gen_finetune}

\paragraph{Metrics} We measure two core metrics for face preservation on average to observe mean change after adapter tuning, as well as per category analysis - \textbf{ID} - stands for face similarity between source and generated person, \textbf{CLIP-T} - measures similarity between image and prompt description in latent space of CLIP model. 

\begin{figure}[!htb]
    \centering
\includegraphics[width=1.0\linewidth]{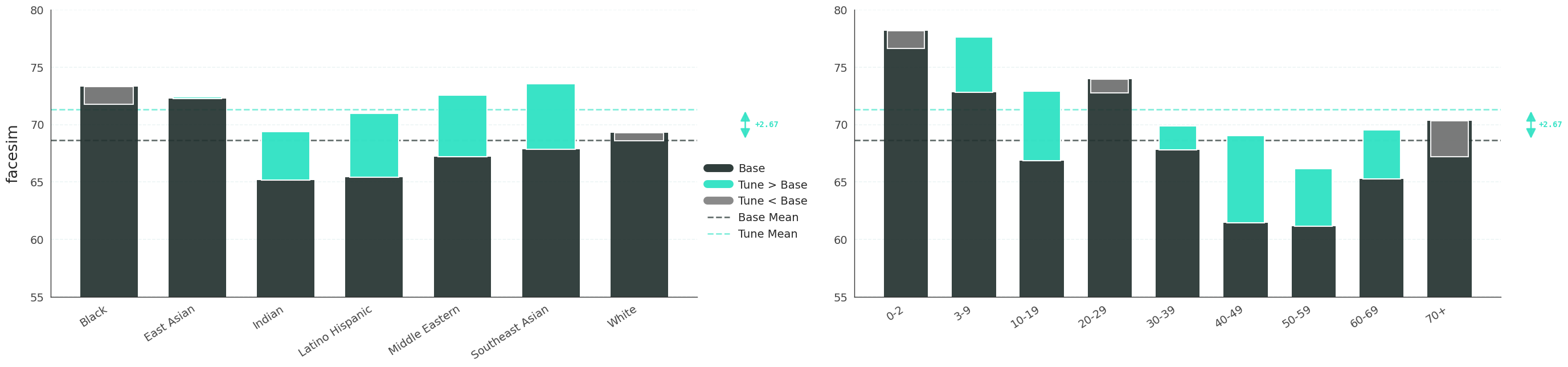}
    \caption{Face similarity measured across different categories of IpAdapter-FaceID before and after tuning on data generated by ID-flow.}
    \label{fig:finetune}
\end{figure}

\paragraph{Evaluation} We use two evaluation benchmarks - one is \textbf{FairBench} introduced before - which is a collection of balanced faces that includes every possible combinations of labels, it's textual captions focus only on face description. This benchmark corresponds of to maximum reconstruction of person facial features. Second benchmark named \textbf{Context} is a less diverse in terms of labels, but includes richer prompts specializing in placing people in contexts, demonstrating trade-off between face similarity being lower for smaller faces and more contextualized humans.

\begin{table}[!hbt]
    \centering
    \caption{Effect of tuning pipelines with our dataset, reporting average metric values}
    \begin{tabular}{lcccc}
        \toprule
        & \multicolumn{2}{c}{\textbf{FairBench}} & \multicolumn{2}{c}{\textbf{Context}} \\
        \cmidrule(lr){2-3} \cmidrule(lr){4-5}
        \textbf{Method} & \textbf{ID} & \textbf{CLIP-T} & \textbf{ID} & \textbf{CLIP-T} \\
        \midrule
        SD-1.5 + FaceID                   & 68.63   & 32.91  & \textbf{28.81}  & \textbf{53.56} \\
        \rowcolor{lightgray}
        SD-1.5 + FaceID (+ tuned)         & \textbf{71.31}   & \textbf{33.27}   & 28.49   & 51.56 \\
        \bottomrule
    \end{tabular}
\end{table}
\vspace{-10pt}

\paragraph{Training details} Finetuning conditional diffusion model comes with a challenge of \textit{data distribution drift}. It's a problem that arises because of discrepancy between data that was used for training original checkpoint and data that is being used for finetuning. Main characteristic that seemingly may hurt performance of these methods in terms of face preservation is learned \textbf{face size}. Face size in collected data serves as hidden factor that can be learned by diffusion model and directly affect similarity metric. Therefore to mitigate effect of this hidden factor we apply approach from \cite{goyal2025preventing} termed \textbf{shortcut finetuning} - specifically we attach frozen ControlNet \cite{zhang2023adding} to condition on human pose and face position in source image, making finetuned IP-adapter layers learn only essential features related to face preservation. Details about this approach and hyperparameters for training are given in Appendix \ref{sec: finetune_details}. 

We provide finetuning results for StableDiffusion coupled with IPAdapter-FaceID. We observe a significant mean improvement in fair face reconstruction benchmark, while keeping comparable performance of contextualized generation. However, it is also important to inspect the change in quality per label. In Fig. \ref{fig:finetune} we showcase improvement across different label values per age and race groups - improvement is not isolated and ultimately makes performance more uniform. More detailed finetuning results with another personalized generation method are provided in Appendix \ref{sec: finetune_details}. 

\subsection{Face Recognition Fairness}
\label{subsec: fr_fairness}

\paragraph{Setup}

We train FR model on a version of our synthetic dataset purposed for this task - we sample 500k images from ID-Flow + Arc2Face with uniform labels, 50 images per identity, and additionally filter identities by similarity threshold in every subgroup.  We compare with two other available FR model baselines: \textbf{SynthFaces-500K} (IResNet-100, ArcFace, $\sim$500K images of 10K identities from our pipeline), \textbf{Vec2Face}~\cite{wu2025vec2face} (IResNet-50 SE, ArcFace, $\sim$300K images, HSFace300K real-web source), and \textbf{HyperFace}~\cite{shahreza2024hyperface}. We evaluate on two regimes: (a)~five standard FR benchmarks (LFW, CFP-FP, AgeDB-30, CALFW, CPLFW) annotated post-hoc with FairFace~\cite{karkkainen2021fairface} demographic labels, and (b)~BFW (Balanced Faces in the Wild)~\cite{robinson2023balancing}, a balanced dataset with over $\sim$900K images balanced across 4 races $\times$ 2 genders with human-assigned labels.

\paragraph{Fairness metrics}

We aim to evaluate fairness of trained FR models. After analyzing standard FR benchmarks we find them to be heavily skewed (AgeDB-30: 88\% White; LFW: 66\% White, 75\% male - see Table~\ref{tab:benchmark_race_dist} in the Appendix), making accuracy-disparity metrics conflate representation with bias. A face verification system accepts or rejects a pair against a threshold $\tau$; the two governing error rates are the \textbf{False Match Rate} (FMR) and \textbf{False Non-Match Rate} (FNMR) \cite{grother2019face}. We evaluate at a calibrated operating point: fix the global FMR to target $\alpha \in \{1\%, 0.1\%\}$ via $\tau_\alpha = \text{Quantile}_\alpha(\{D_i : y_i{=}0\})$, then measure per-group FNMR. This reflects realistic deployment where an operator fixes a tolerable false-alarm rate. To normalize for absolute model quality and isolate \emph{relative} group (dis)advantage, we adopt the \textbf{Skewed Error Ratio} (SER), inspired by \cite{wang2020mitigating}:
\begin{equation}
    \text{SER}_g(\alpha) = \frac{\text{FNMR}_g(\alpha)}{\text{FNMR}_{\text{global}}(\alpha)}.
    \label{eq:ser}
\end{equation}
Unlike the original formulation - which takes the ratio of worst to best group — we anchor to the global population FNMR, avoiding division-by-zero when any group achieves perfect recall and decoupling group disadvantage from the best group's advantage. $\text{SER}=1.0$ is perfect parity; $\text{SER}>1$ indicates disproportionate miss burden. We report $\text{SER}_{\max}=\max_g \text{SER}_g$ (lower\,=\,fairer).

\begin{wraptable}{!t}{0.6\linewidth}
    \vspace{-1.0\baselineskip}
    \caption{BFW fairness aggregate ($\downarrow$\,fairer). \textbf{Bold}\,=\,best.}
    \label{tab:bfw_aggregate}
    \resizebox{\linewidth}{!}{
    \begin{tabular}{lccc}
        \toprule
        \textbf{Model} & Acc Disp\,$\downarrow$ & SER\textsubscript{max}@1\%\,$\downarrow$ & SER\textsubscript{max}@0.1\%\,$\downarrow$ \\
        \midrule
        Vec2Face         & 0.087          & 1.107          & 1.057          \\
        HyperFace        & 0.068          & 1.143          & 1.101          \\
        \textbf{SynthFaces-500K (ours)} & \textbf{0.043} & \textbf{1.074} & \textbf{1.011} \\
        \bottomrule
    \end{tabular}
    }
\end{wraptable}

\paragraph{Results on standard benchmarks}

Table~\ref{tab:ser_benchmarks} reports the maximum racial SER at the tighter FPR\,=\,0.1\% operating point. SynthFaces-500K achieves the lowest $\text{SER}_{\max}$ on 4 of 5 benchmarks and the best mean across all five. The 40\% reduction in mean racial SER relative to Vec2Face (1.097 vs.\ 1.531) indicates that a model trained on our demographically balanced data imposes substantially less differential miss burden on minority groups at high-precision operating points. Additionally accuracy figures are provided in Table~\ref{tab:global_acc_fr} in the Appendix, and we also report FPR\,=\,1\% SER results in Table~\ref{tab:ser_fpr1}; SynthFaces-500K's lower overall accuracy reflects that a model trained on a more balanced dataset may still underperform in terms of global accuracy reported on established benchmarks.

\begin{table}[!htb]
    \centering
    \caption{Maximum racial SER at FPR\,=\,0.1\% on standard FR benchmarks ($\downarrow$ fairer, 1.0\,=\,perfect parity). \textbf{Bold}\,=\,best. $\star$\,=\,SynthFaces-500K is fairest.}
    \label{tab:ser_benchmarks}
    \resizebox{\textwidth}{!}{
    \begin{tabular}{lcccccr}
        \toprule
        \textbf{Model} & LFW & CFP-FP & AgeDB-30 & CALFW & CPLFW & \textbf{Mean} \\
        \midrule
        Vec2Face-HSFace300K   & 2.217 & 1.296              & 1.560              & 1.185 & 1.397              & 1.531 \\
        HyperFace-10K-LDM     & 2.005 & 1.096              & 1.176              & \textbf{1.041} & 1.085 & 1.280 \\
        \textbf{SynthFaces-500K (ours)} & \textbf{1.197} & \textbf{1.062} & \textbf{1.054} & 1.170 & \textbf{1.003} & \textbf{1.097} \\
        \bottomrule
    \end{tabular}
    }
\end{table}

\paragraph{Results on BFW}

Table~\ref{tab:bfw_aggregate} reports aggregate fairness metrics on the balanced BFW benchmark. SynthFaces-500K achieves the lowest accuracy disparity (0.043 vs.\ 0.087 for Vec2Face, 0.068 for HyperFace), the lowest $\text{SER}_{\max}$ at both operating points, and near-perfect parity at FPR\,=\,0.1\% ($\text{SER}_{\max}=1.011$); SynthFaces-500K consistently reduces disparity.

\subsection{Downstream VQA: Demographic Attribute Classification}
\label{sec:vqa_fairness}

\paragraph{Setup}

We fine-tune \textbf{LLaVA-1.5-7B}~\cite{liu2023visual} with LoRA adapters on demographic VQA data from three sources: (a)~\textbf{SynthFaces-HQ} - samples from our datasets paired with multiple-choice questions on gender, race, and age derived from pipeline annotations; (b)~\textbf{SynthFaces} - mirrors HQ version, but uses non-HQ images with just human faces (c)~\textbf{Vec2Face} - a VQA set from arc2face-rendered images annotated with FairFace labels. All runs follow an identical LoRA fine-tuning protocol; checkpoints are selected by fairness/accuracy trade-off on a held-out split.

\paragraph{Metrics}

For tasks $T$ and $S$ - predictive and sensetive attributes, $T \neq S \in \{\text{gender}, \text{race}, \text{age}\}$, we define the \textbf{cross-attribute accuracy disparity}:
\begin{equation}
    \Delta(T,S) = \max_{s \in \mathcal{S}} \mathrm{Acc}(T \mid S{=}s) - \min_{s \in \mathcal{S}} \mathrm{Acc}(T \mid S{=}s).
    \label{eq:delta_ts}
\end{equation}
$\Delta(T,S)=0$ is perfect cross-attribute fairness. We average over all 6 ordered pairs to obtain the \textbf{average disparity} $\overline{\Delta}$ (lower\,=\,fairer).

\paragraph{Evaluation}

We evaluate on a held-out internal VQA benchmark (37,316 questions). To avoid penalising either model for a domain gap, each fine-tuned model is evaluated on images from its \emph{own} training domain, with a domain-matched untuned LLaVA baseline in each row.

\begin{wraptable}{r}{0.6\linewidth}
    \caption{Domain-aligned VQA results Avg acc\,=\,mean over gender/race/age ($\uparrow$); $\overline{\Delta}\,\downarrow$ fairer. \textbf{Bold}\,=\,best per domain.}
    \label{tab:vqa_results}
    \resizebox{\linewidth}{!}{
    \begin{tabular}{llcc}
        \toprule
        \textbf{Domain} & \textbf{Model} & Avg acc\,$\uparrow$ & $\overline{\Delta}$\,$\downarrow$ \\
        \midrule
        HQ     & Baseline             & 0.551          & 0.183 \\
        HQ     & \textbf{SynthFaces-HQ (ours)} & \textbf{0.700} & \textbf{0.151} \\
        \midrule
        non-HQ & Baseline             & 0.645          & 0.176 \\
        non-HQ & Vec2Face             & 0.559          & 0.263 \\
        non-HQ & \textbf{SynthFaces-nonHQ (ours)} & \textbf{0.797} & \textbf{0.138} \\
        \bottomrule
    \end{tabular}
    }
\end{wraptable}

Both SynthFaces variants improve over their domain baselines: \textbf{SynthFaces-HQ} raises avg accuracy from 0.551 to \textbf{0.700} ($\overline{\Delta}$ 0.183$\to$\textbf{0.151}); \textbf{SynthFaces-nonHQ} - trained on arc2face-style renders of the same identities - achieves the strongest result in this domain-aligned evaluation (avg acc \textbf{0.797}, $\overline{\Delta}$ \textbf{0.138}), including $+$23.3\,pp race accuracy over baseline. The domain-controlled comparison isolates training data composition: \textbf{SynthFaces-nonHQ vs Vec2Face} use identical image style and resolution, yet SynthFaces achieves $+$0.125 lower $\overline{\Delta}$ and $+$23.8\,pp higher avg accuracy, consistent with its more balanced identity distribution. Vec2Face fine-tuning, by contrast, regresses on gender and age, suggesting its narrower training distribution degrades the pre-trained model's demographic priors.

\section{Conclusion}

We introduced SynthFaces, a large-scale, identity-grouped and captioned synthetic human dataset designed for demographic balance while reducing reliance on released images of real individuals. Its ID-Flow sampler provides explicit demographic control while maintaining identity diversity and empirical separation from the real identities used for training. Across personalized generation, face recognition, and demographic VQA, models trained with SynthFaces improve subgroup parity while retaining useful task performance. These results show that controlling identity composition during dataset construction is an effective lever for fairer human-centric models. Future work should reduce dependence on potentially biased face encoders and foundation generators and investigate hybrid real--synthetic balancing strategies.

\section{Limitations}

While SynthFaces addresses key gaps in existing datasets, several limitations should be acknowledged.

\paragraph{Bias inheritance from foundation models.} Arc2Face and Flux despite being state of the art generators for corresponding parts of pipeline carry aesthetic and demographic biases from their respective training corpora. Our pipeline guarantees label-level balance across age, race, and gender, but within-group visual diversity and image style remain bounded by what these upstream models can faithfully generate. Demographic combinations that are underrepresented in Arc2Face's training data may yield lower-fidelity or less varied outputs.

\paragraph{Synthetic-to-real distribution gap.} Images produced by diffusion models may contain subtle distributional artifacts absent from real photographs. While our VLM-based filtering step removes overt defects, residual domain differences could limit transferability when SynthFaces-trained models are evaluated on real-world data.

\paragraph{Dependency on upstream model capabilities.} The expressive range of generated identities is ultimately constrained by Arc2Face's capacity for identity preservation and Flux's generation quality. Improvements to either foundation model would directly benefit downstream dataset quality, but also introduce a moving dependency.

\section{Ethics statement} 

The creation of SynthFaces is a direct response to the ethical crisis surrounding web-scraped human-centric datasets, many of which have been rightfully retracted due to non-consensual use of personal images, including those of minors. To reduce these privacy and consent risks, SynthFaces releases synthetic identities generated by our pipeline rather than scraped personal images. Because ID-Flow is trained from ArcFace embeddings extracted from real datasets, we do not claim a formal privacy guarantee; instead, we empirically evaluate nearest-neighbor separation with the ID-NonLeak metric and design the dataset to avoid direct re-use of real personal images.

Furthermore, we address the issue of demographic bias that plagues both real and synthetic datasets. Our pipeline is explicitly designed to generate a balanced representation across age, gender, and ethnicity, creating a resource purposed for training fairer models and mitigating the harmful disparities common in existing technologies. While we acknowledge that no synthetic data is perfect and subtle biases may persist, we believe this work demonstrates a responsible path forward, decoupling progress in human-centric computer vision from the exploitation of personal data.

\section{Acknowledgment}

We gratefully acknowledge HSE University for providing access to its computational cluster, which was used to conduct the experiments presented in this work. Support from the Basic Research Program of HSE University is gratefully acknowledged (HSE-BR-2025-019).

\bibliography{main}

@book{grother2019face,
  title={Face recognition vendor test (fvrt): Part 3, demographic effects},
  author={Grother, Patrick and Ngan, Mei and Hanaoka, Kayee},
  year={2019},
  publisher={National Institute of Standards and Technology Gaithersburg, MD}
}

@article{liu2023visual,
  title={Visual instruction tuning},
  author={Liu, Haotian and Li, Chunyuan and Wu, Qingyang and Lee, Yong Jae},
  journal={Advances in neural information processing systems},
  volume={36},
  pages={34892--34916},
  year={2023}
}

@inproceedings{wang2020mitigating,
  title={Mitigating bias in face recognition using skewness-aware reinforcement learning},
  author={Wang, Mei and Deng, Weihong},
  booktitle={Proceedings of the IEEE/CVF conference on computer vision and pattern recognition},
  pages={9322--9331},
  year={2020}
}

@article{robinson2023balancing,
  title={Balancing biases and preserving privacy on balanced faces in the wild},
  author={Robinson, Joseph P and Qin, Can and Henon, Yann and Timoner, Samson and Fu, Yun},
  journal={IEEE Transactions on Image Processing},
  volume={32},
  pages={4365--4377},
  year={2023},
  publisher={IEEE}
}

@inproceedings{wu2025vec2face,
  title={Vec2face: Scaling face dataset generation with loosely constrained vectors},
  author={Wu, Haiyu and Singh, Jaskirat and Tian, Sicong and Zheng, Liang and Bowyer, Kevin},
  booktitle={International Conference on Learning Representations},
  volume={2025},
  pages={23490--23503},
  year={2025}
}

@inproceedings{ruiz2023dreambooth,
  title={Dreambooth: Fine tuning text-to-image diffusion models for subject-driven generation},
  author={Ruiz, Nataniel and Li, Yuanzhen and Jampani, Varun and Pritch, Yael and Rubinstein, Michael and Aberman, Kfir},
  booktitle={Proceedings of the IEEE/CVF conference on computer vision and pattern recognition},
  pages={22500--22510},
  year={2023}
}

@article{shahreza2024hyperface,
  title={HyperFace: Generating synthetic face recognition datasets by exploring face embedding hypersphere},
  author={Shahreza, Hatef Otroshi and Marcel, S{\'e}bastien},
  journal={arXiv preprint arXiv:2411.08470},
  year={2024}
}

@article{kynkaanniemi2019improved,
  title={Improved precision and recall metric for assessing generative models},
  author={Kynk{\"a}{\"a}nniemi, Tuomas and Karras, Tero and Laine, Samuli and Lehtinen, Jaakko and Aila, Timo},
  journal={Advances in neural information processing systems},
  volume={32},
  year={2019}
}

@article{ho2022classifier,
  title={Classifier-free diffusion guidance},
  author={Ho, Jonathan and Salimans, Tim},
  journal={arXiv preprint arXiv:2207.12598},
  year={2022}
}

@inproceedings{rombach2022high,
  title={High-resolution image synthesis with latent diffusion models},
  author={Rombach, Robin and Blattmann, Andreas and Lorenz, Dominik and Esser, Patrick and Ommer, Bj{\"o}rn},
  booktitle={Proceedings of the IEEE/CVF conference on computer vision and pattern recognition},
  pages={10684--10695},
  year={2022}
}

@inproceedings{kemelmacher2016megaface,
  title={The megaface benchmark: 1 million faces for recognition at scale},
  author={Kemelmacher-Shlizerman, Ira and Seitz, Steven M and Miller, Daniel and Brossard, Evan},
  booktitle={Proceedings of the IEEE conference on computer vision and pattern recognition},
  pages={4873--4882},
  year={2016}
}

@inproceedings{ismayilov2025fluxsynid,
  title={FLUXSynID: A Framework for Identity-Controlled Synthetic Face Generation with Document and Live Images},
  author={Ismayilov, Raul and Sero, Dzemila and Spreeuwers, Luuk},
  booktitle={Proceedings of the IEEE/CVF International Conference on Computer Vision},
  pages={3698--3708},
  year={2025}
}

@inproceedings{zhang2023adding,
  title={Adding conditional control to text-to-image diffusion models},
  author={Zhang, Lvmin and Rao, Anyi and Agrawala, Maneesh},
  booktitle={Proceedings of the IEEE/CVF international conference on computer vision},
  pages={3836--3847},
  year={2023}
}

@article{dai2025humanvlm,
  title={Humanvlm: Foundation for human-scene vision-language model},
  author={Dai, Dawei and Xu, Long and Li, Yutang and Zhang, Yuanhui and Xia, Shuyin},
  journal={Information Fusion},
  volume={123},
  pages={103271},
  year={2025},
  publisher={Elsevier}
}

@article{goyal2025preventing,
  title={Preventing shortcuts in adapter training via providing the shortcuts},
  author={Goyal, Anujraaj Argo and Qian, Guocheng Gordon and Coskun, Huseyin and Gupta, Aarush and Tam, Himmy and Ostashev, Daniil and Hu, Ju and Sagar, Dhritiman and Tulyakov, Sergey and Aberman, Kfir and others},
  journal={arXiv preprint arXiv:2510.20887},
  year={2025}
}

@inproceedings{guo2016ms,
  title={Ms-celeb-1m: A dataset and benchmark for large-scale face recognition},
  author={Guo, Yandong and Zhang, Lei and Hu, Yuxiao and He, Xiaodong and Gao, Jianfeng},
  booktitle={European conference on computer vision},
  pages={87--102},
  year={2016},
  organization={Springer}
}

@article{li2023survey,
  title={A survey on fairness in large language models},
  author={Li, Yingji and Du, Mengnan and Song, Rui and Wang, Xin and Wang, Ying},
  journal={arXiv preprint arXiv:2308.10149},
  year={2023}
}

@article{li2024id,
  title={ID$^3$: Identity-Preserving-yet-Diversified Diffusion Models for Synthetic Face Recognition},
  author={Li, Shen and Xu, Jianqing and Wu, Jiaying and Xiong, Miao and Deng, Ailin and Ji, Jiazhen and Huang, Yuge and Feng, Wenjie and Ding, Shouhong and Hooi, Bryan},
  journal={arXiv preprint arXiv:2409.17576},
  year={2024}
}

@inproceedings{li2024photomaker,
  title={Photomaker: Customizing realistic human photos via stacked id embedding},
  author={Li, Zhen and Cao, Mingdeng and Wang, Xintao and Qi, Zhongang and Cheng, Ming-Ming and Shan, Ying},
  booktitle={Proceedings of the IEEE/CVF conference on computer vision and pattern recognition},
  pages={8640--8650},
  year={2024}
}

@inproceedings{melzi2023gandiffface,
  title={Gandiffface: Controllable generation of synthetic datasets for face recognition with realistic variations},
  author={Melzi, Pietro and Rathgeb, Christian and Tolosana, Ruben and Vera-Rodriguez, Ruben and Lawatsch, Dominik and Domin, Florian and Schaubert, Maxim},
  booktitle={Proceedings of the IEEE/CVF International Conference on Computer Vision},
  pages={3086--3095},
  year={2023}
}

@inproceedings{karkkainen2021fairface,
  title={Fairface: Face attribute dataset for balanced race, gender, and age for bias measurement and mitigation},
  author={Karkkainen, Kimmo and Joo, Jungseock},
  booktitle={Proceedings of the IEEE/CVF winter conference on applications of computer vision},
  pages={1548--1558},
  year={2021}
}

@article{ye2023ip,
  title={Ip-adapter: Text compatible image prompt adapter for text-to-image diffusion models},
  author={Ye, Hu and Zhang, Jun and Liu, Sibo and Han, Xiao and Yang, Wei},
  journal={arXiv preprint arXiv:2308.06721},
  year={2023}
}

@article{wang2024instantid,
  title={Instantid: Zero-shot identity-preserving generation in seconds},
  author={Wang, Qixun and Bai, Xu and Wang, Haofan and Qin, Zekui and Chen, Anthony and Li, Huaxia and Tang, Xu and Hu, Yao},
  journal={arXiv preprint arXiv:2401.07519},
  year={2024}
}

@inproceedings{papantoniou2024arc2face,
  title={Arc2face: A foundation model for id-consistent human faces},
  author={Papantoniou, Foivos Paraperas and Lattas, Alexandros and Moschoglou, Stylianos and Deng, Jiankang and Kainz, Bernhard and Zafeiriou, Stefanos},
  booktitle={European Conference on Computer Vision},
  pages={241--261},
  year={2024},
  organization={Springer}
}

@article{mehrabi2021survey,
  title={A survey on bias and fairness in machine learning},
  author={Mehrabi, Ninareh and Morstatter, Fred and Saxena, Nripsuta and Lerman, Kristina and Galstyan, Aram},
  journal={ACM computing surveys (CSUR)},
  volume={54},
  number={6},
  pages={1--35},
  year={2021},
  publisher={ACM New York, NY, USA}
}

@article{lipman2024flow,
  title={Flow matching guide and code},
  author={Lipman, Yaron and Havasi, Marton and Holderrieth, Peter and Shaul, Neta and Le, Matt and Karrer, Brian and Chen, Ricky TQ and Lopez-Paz, David and Ben-Hamu, Heli and Gat, Itai},
  journal={arXiv preprint arXiv:2412.06264},
  year={2024}
}

@inproceedings{fu2022stylegan,
  title={Stylegan-human: A data-centric odyssey of human generation},
  author={Fu, Jianglin and Li, Shikai and Jiang, Yuming and Lin, Kwan-Yee and Qian, Chen and Loy, Chen Change and Wu, Wayne and Liu, Ziwei},
  booktitle={European Conference on Computer Vision},
  pages={1--19},
  year={2022},
  organization={Springer}
}

@article{kvanchiani2023easyportrait,
  title={EasyPortrait--Face Parsing and Portrait Segmentation Dataset},
  author={Kvanchiani, Karina and Petrova, Elizaveta and Efremyan, Karen and Sautin, Alexander and Kapitanov, Alexander},
  journal={arXiv preprint arXiv:2304.13509},
  year={2023}
}

@inproceedings{li2024cosmicman,
  title={Cosmicman: A text-to-image foundation model for humans},
  author={Li, Shikai and Fu, Jianglin and Liu, Kaiyuan and Wang, Wentao and Lin, Kwan-Yee and Wu, Wayne},
  booktitle={Proceedings of the IEEE/CVF Conference on Computer Vision and Pattern Recognition},
  pages={6955--6965},
  year={2024}
}

@inproceedings{kim2023dcface,
  title={Dcface: Synthetic face generation with dual condition diffusion model},
  author={Kim, Minchul and Liu, Feng and Jain, Anil and Liu, Xiaoming},
  booktitle={Proceedings of the ieee/cvf conference on computer vision and pattern recognition},
  pages={12715--12725},
  year={2023}
}

@inproceedings{bae2023digiface,
  title={Digiface-1m: 1 million digital face images for face recognition},
  author={Bae, Gwangbin and de La Gorce, Martin and Baltru{\v{s}}aitis, Tadas and Hewitt, Charlie and Chen, Dong and Valentin, Julien and Cipolla, Roberto and Shen, Jingjing},
  booktitle={Proceedings of the IEEE/CVF Winter Conference on Applications of Computer Vision},
  pages={3526--3535},
  year={2023}
}

@article{dosi2025harmonizing,
  title={Harmonizing geometry and uncertainty: Diffusion with hyperspheres},
  author={Dosi, Muskan and Chiranjeev, Chiranjeev and Thakral, Kartik and Vatsa, Mayank and Singh, Richa},
  journal={arXiv preprint arXiv:2506.10576},
  year={2025}
}

@article{guo2024pulid,
  title={Pulid: Pure and lightning id customization via contrastive alignment},
  author={Guo, Zinan and Wu, Yanze and Zhuowei, Chen and Zhang, Peng and He, Qian and others},
  journal={Advances in neural information processing systems},
  volume={37},
  pages={36777--36804},
  year={2024}
}

@article{xu2023imagereward,
  title={Imagereward: Learning and evaluating human preferences for text-to-image generation},
  author={Xu, Jiazheng and Liu, Xiao and Wu, Yuchen and Tong, Yuxuan and Li, Qinkai and Ding, Ming and Tang, Jie and Dong, Yuxiao},
  journal={Advances in Neural Information Processing Systems},
  volume={36},
  pages={15903--15935},
  year={2023}
}

@inproceedings{jiang2025infiniteyou,
  title={Infiniteyou: Flexible photo recrafting while preserving your identity},
  author={Jiang, Liming and Yan, Qing and Jia, Yumin and Liu, Zichuan and Kang, Hao and Lu, Xin},
  booktitle={Proceedings of the IEEE/CVF International Conference on Computer Vision},
  pages={10898--10907},
  year={2025}
}

@article{wu2025qwen,
  title={Qwen-image technical report},
  author={Wu, Chenfei and Li, Jiahao and Zhou, Jingren and Lin, Junyang and Gao, Kaiyuan and Yan, Kun and Yin, Sheng-ming and Bai, Shuai and Xu, Xiao and Chen, Yilei and others},
  journal={arXiv preprint arXiv:2508.02324},
  year={2025}
}

@misc{flux-2-2025,
    author={Black Forest Labs},
    title={{FLUX.2: Frontier Visual Intelligence}},
    year={2025},
    howpublished={\url{https://bfl.ai/blog/flux-2}},
}

\newpage
\appendix
\section{Appendix}




\subsection{Detailed datasets comparison}
\label{sec: detail_data_compare}

In Table \ref{tab: datasets-details} we provide a general overview of existing datasets in the field that are potentially suitable for various tasks involving images with humans. For our dataset the usage license is in the process of development, as we aim to make sure that synthetic data is still protected from application in unethical and harmful domains.

\begin{sidewaystable}[!htbp]
    \centering
    \caption{Detailed comparison of human image datasets purposed for various tasks}
    \label{tab: datasets-details}
    \resizebox{\textwidth}{!}{
    \begin{tabular}{l c c c c c c c c c}
        \toprule
        \textbf{Name} & \textbf{Task} & \textbf{Size} & \textbf{Avg. resolution} & \textbf{Captions} & \textbf{Human w/ context} & \textbf{ID-grouped} & \textbf{Data Source} & \textbf{License} \\
        \midrule
        CosmicMan & text-driven generation & 5.5M & 1488×1255 & yes & yes & no & real & NON COMMERCIAL \\
        Human-Art & detection, generation & 50k & 1115×1287 & yes & yes & no & real \& synthetic & CC, NON COMMERCIAL \\
        DF-MM & text-driven generation & 44k & 750 × 1101 & yes & no & no & real & NON COMMERCIAL \\
        LAION-Human & generation & 1M & 688 × 650 & yes & yes & no & real & CC BY 4.0 \\
        SHHQ & generation & 40k & 1024 × 512 & no & no & no & real & CC0 \\
        EasyPortrait & segmentation & 40k & HD & no & no & no & real & custom \\
        VGGFace2 & facerecog, generation & 3.31M & LowRes & no & no & yes & real & MIT \\
        MultiID-2M & multiperson-gen & 1M & HD & yes & yes & no & real & Apache \\
        FFHQ/\newline FFHQ-in-the-wild & generation (faces) & 70k & HD & no & no/yes & no & real & CC \\
        \midrule
        DCFace & facerecog & 500k & LowRes & no & yes & yes & synthetic & ? \\
        VariFace & facerecog & 500k & LowRes & no & yes & yes & synthetic & ? \\
        ID$^3$ & facerecog & 500k & LowRes & no & yes & yes & synthetic & ? \\
        HyperSphere & facerecog & 500k & LowRes & no & yes & yes & synthetic & CC \\
        Synthetics-DisCo & facerecog  & LowRes & 650k & no & yes & yes & synthetic & CC, NON COMMERCIAL \\
        FluxSynID & facerecog & 50k & HD & yes & no & no & synthetic & GNU \\
        \midrule
        SynthFaces (ours) & persongen & 150k & HD & yes & yes & yes & synthetic & \textbf{in process} \\
        \bottomrule
    \end{tabular}
    }
\end{sidewaystable}

\subsection{Definitions and details of flow matching on sphere}
\label{sec: sphere projections}

\subsubsection{Riemannian Manifolds and the Sphere}

A smooth manifold $\mathcal{M}$ of dimension $d$ is a topological space that locally resembles Euclidean space $\mathbb{R}^d$, though globally it may have curvature. At each point $x \in \mathcal{M}$, the tangent space $\mathcal{T}_x\mathcal{M}$ consists of all vectors tangent to $\mathcal{M}$ at $x$. The collection of all tangent spaces forms the tangent bundle $\mathcal{T}\mathcal{M} = \bigcup_{x\in\mathcal{M}} \mathcal{T}_x\mathcal{M}$. A Riemannian manifold $(\mathcal{M}, g)$ is equipped with an inner product $g_x(\cdot,\cdot)$ on each tangent space, enabling the measurement of vector norms and geodesic distances. In this work, we consider the $n$-dimensional unit sphere $\mathbb{S}^{n-1} = \{x \in \mathbb{R}^{n} : \|x\| = 1\}$ embedded in $\mathbb{R}^{n}$, with the Riemannian metric inherited from the Euclidean inner product.

For the sphere, two fundamental maps admit closed-form expressions. The exponential map $\exp_x : \mathcal{T}_x\mathbb{S}^{n-1} \to \mathbb{S}^{n-1}$ takes a tangent vector $v \in \mathcal{T}_x\mathbb{S}^{n-1}$ and returns the point reached by following the geodesic starting at $x$ with initial velocity $v$ for unit time:
\begin{equation}
\exp_x(v) = \cos(\|v\|) x + \sin(\|v\|) \frac{v}{\|v\|}.
\end{equation}
Conversely, the logarithmic map $\log_x : \mathbb{S}^{n-1} \to \mathcal{T}_x\mathbb{S}^{n-1}$ returns the unique tangent vector $v$ such that $\exp_x(v) = y$, with norm equal to the geodesic distance:
\begin{equation}
\log_x(y) = \frac{\theta}{\sin\theta}(y - \langle x,y\rangle x),\quad \theta = \arccos(\langle x,y\rangle),
\end{equation}
where $\langle\cdot,\cdot\rangle$ denotes the standard Euclidean inner product. These maps provide the necessary tools for constructing geodesic interpolations directly on the manifold.

\subsubsection{Geodesic Conditional Flows and Riemannian Flow Matching}

Flow matching (FM) is a simulation-free framework for training continuous normalizing flows. In Euclidean space, one defines a conditional probability path $p_t(x|x_1)$ that interpolates between a simple prior $p_0 = p$ and a data point $x_1$, typically via the linear interpolation $x_t = t x_1 + (1-t)x_0$ with $x_0 \sim p$. The corresponding conditional velocity field is $u_t(x|x_1) = x_1 - x_0$. Training then proceeds by regressing a neural network $v_t^\theta(x)$ to match these conditional velocities via the conditional flow matching (CFM) loss $\mathcal{L}_{\mathrm{CFM}} = \mathbb{E}_{t,x_0,x_1}\|v_t^\theta(x_t) - (x_1 - x_0)\|^2$.

For Riemannian manifolds, the linear interpolation is replaced by geodesic interpolation. Following \cite{lipman2024flow}, we define the \textit{geodesic conditional flow}:
\begin{equation}
\phi_t(x_0|x_1) = \exp_{x_0}(\kappa(t) \log_{x_0}(x_1)),\quad t\in[0,1],
\end{equation}
where $\kappa:[0,1]\to[0,1]$ is a smooth scheduler satisfying $\kappa(0)=0$ and $\kappa(1)=1$. This flow moves $x_0$ along the geodesic toward $x_1$ at constant speed when $\kappa(t)=t$. The conditional probability path is then $p_t(x|x_1) = [\phi_t(\cdot|x_1)]_\sharp p(x_0)$, the push-forward of the prior along the geodesic. The corresponding conditional velocity field is obtained by differentiation ($D$ denotes 'differential' of map):
\begin{equation}
u_t(x_t|x_1) = \dot{\phi}_t(\phi_t^{-1}(x_t|x_1)|x_1) = \dot{\kappa}(t) D\exp_{x_0}(\kappa(t)\log_{x_0}(x_1))[\log_{x_0}(x_1)],
\end{equation}
which for the scheduler $\kappa(t)=t$ simplifies to:
\begin{equation}
u_t(x_t|x_1) = -D\exp_{x_1}((1-t)\log_{x_1}(x_0))[\log_{x_1}(x_0)].
\end{equation}
Importantly, for the sphere these quantities are available in closed form via the exponential and logarithmic maps, enabling simulation-free training.

\subsubsection{Training Objective and Marginal Velocity}

The Riemannian conditional flow matching (RCFM) loss trains a neural vector field $v_t^\theta(x) \in \mathcal{T}_x\mathcal{M}$ to match the conditional velocity:

\begin{equation}
\mathcal{L}_{\mathrm{RCFM}}(\theta) = \mathbb{E}_{t\sim\mathcal{U}[0,1], x_0\sim p, x_1\sim p_{\mathrm{data}}} \|v_t^\theta(x_t) - u_t(x_t|x_1)\|^2_g,
\end{equation}

where $\|\cdot\|_g$ is the Riemannian norm on $\mathcal{T}_{x_t}\mathcal{M}$ and $x_t = \phi_t(x_0|x_1)$. The marginal probability path $p_t(x) = \int p_t(x|x_1)p_{\mathrm{data}}(x_1)\,\mathrm{dvol}_{x_1}$ is generated by marginal velocity field given by the conditional expectation:
\begin{equation}
u_t(x) = \mathbb{E}_{x_1\sim p_{1|t}(\cdot|x)}[u_t(x|x_1)],
\end{equation}
where $p_{1|t}(x_1|x) = p_t(x|x_1)p_{\mathrm{data}}(x_1)/p_t(x)$ is the posterior. By the marginalization trick (Theorem 10, \cite{lipman2024flow}), the minimizer of the RCFM loss indeed recovers $u_t(x)$. At inference, samples are generated by drawing $x_0\sim p$ (e.g., uniform on the sphere) and integrating the learned vector field $\dot{x}_t = v_t^\theta(x_t)$ from $t=0$ to $t=1$ using a Riemannian ODE solver. For the sphere, integration can be performed by projecting updates onto the tangent space and mapping back via the exponential map at each step.




\subsection{Additional data and model details}
\label{sec:model_details_and_ablations}

\begin{figure}[bth!]
    \centering
\includegraphics[width=1.0\linewidth]{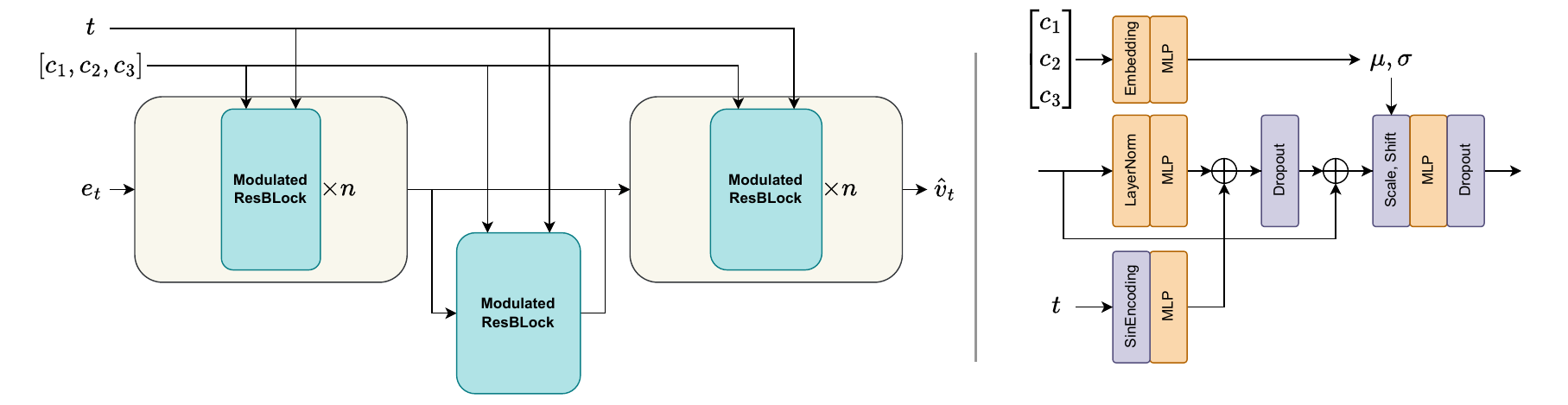}
    \caption{Left - high level architecture of the ID-flow model, with conceptual parts corresponding to encoder/decoder architecture, Right - architecture of \texttt{Modulated ResBlock}, which consists of parts inspired by DiT architectures, but without attention mechanisms, orange color highlights trainable parameters}
    \label{fig:model_architecture}
\end{figure}

\paragraph{Data collection} Our training stage is conducted fully in latent space of ArcFace embedding. To obtain them, we first processed images from Cosmic-HQ, FFHQ, EasyPortait and also enriched with images of people from the internet. We included only embeddings that were detectable with fixed bounding box size set to 640, since embeddings tend to significantly differ depending on detection size. We implement a deduplication stage, where all collected embeddings are filtered out by a threshold value of 0.4 pairwise cosine similarity, ensuring diverse faces are used for training and avoiding identity collapse, where model learns only fixed amount of identities. Then we also annotate all collected images with FairFace, providing age/race/gender labels, and remove embeddings that correspond to images that failed labeling stage. As a result, we train on around 50k identity embeddings with labels. For evaluation we use hold-out set of Cosmic-HQ images and CelebA, which weren't used during training, and apply same pipeline to obtain embeddings and labels, without deduplication stage.

\paragraph{Model details and ablations} Model consists mainly of several MLP layers with intermediate activations combined with LayerNorm, Dropout and modulation layers to condition on labels - visual diagram of architecture is given in Figure \ref{fig:model_architecture}. Since we encode 3 independent labels for each category, corresponding layers that predict values for scale and shift operation have $\times 3$ dimensionality output size. Ablated model parameters are given in Table \ref{tab:model_ablation}.

\begin{table}[!htb]
    \centering
    \caption{Ablated parameters of model}
    \label{tab:model_ablation}
    \resizebox{\textwidth}{!}{
    \begin{tabular}{ccccccc}
        \toprule
        block\_num & hidden\_dim & c\_embed dim & attention & activation & optimizer & schedule \\
        \midrule
        \begin{tabular}{@{}c@{}} 2, \\ 4, \\ 6 \end{tabular} & \begin{tabular}{@{}c@{}} 128, \\ 256, \\ 512 \end{tabular} & \begin{tabular}{@{}c@{}} 128, \\ 256, \\ 512, \end{tabular} & \begin{tabular}{@{}c@{}} yes, \\ no \end{tabular} & \begin{tabular}{@{}c@{}} LeakyRelu \\ Sigmoid \\ SoftSign \\ GeLu  \end{tabular} &  \begin{tabular}{@{}c@{}} SGD, \\ Adam, \\ AdamW \end{tabular} &  \begin{tabular}{@{}c@{}} const, \\ linear warmup \end{tabular} \\ 
        \bottomrule
    \end{tabular}
    }
\end{table}

\begin{figure}[bth!]
    \centering
\includegraphics[scale=0.5]{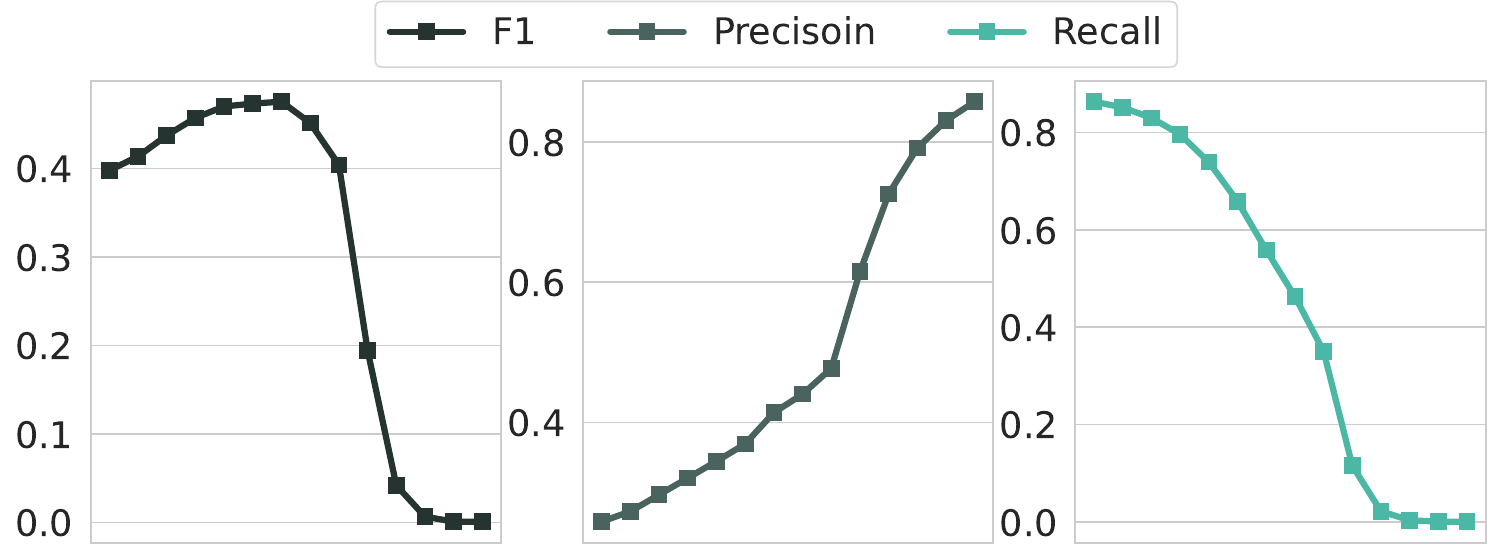}
    \caption{Trade-off of inference quality of the model embedding samples between fidelity and variability of the samples. We change guidance scale values in range from 1.0 to 8.0}
    \label{fig:gs_ablation}
\end{figure}

In the end we choose configuration of \texttt{block\_num} = 2 (meaning 2 blocks in encoder and decoder parts), \texttt{hidden\_dim}=512, \texttt{c\_embed\_dim}=128, \texttt{GeLU} activations. We removed attention blocks, as even with minimal amount of attention heads model began to overfit and collapsed into few identities. Resulting lightweight model has around 11M parameters, was trained for 1000 epochs with Adam optimizer with constant learning rate set to 0.0003. We applied small inter-batch slerp interpolation between embeddings as augmentation technique. Total training took around 3.5 hours on a single V100 GPU. 

\paragraph{Inference trade-off} Our ID-flow model offers a superior trade-off compared to other baselines in terms of embedding fidelity/variance. However this can be directly controlled by playing with guidance scale parameter of identity-flow model. In Fig. \ref{fig:gs_ablation} we show resulting Pareto front for F1, Precision and Recall - model transitions from variability/low-quality to mode collapse behavior. 

\subsection{Similarity filtering of identities}
\label{sec: sim_filter_details}

\begin{figure}[!bht]
    \centering
    \includegraphics[width=0.95\linewidth]{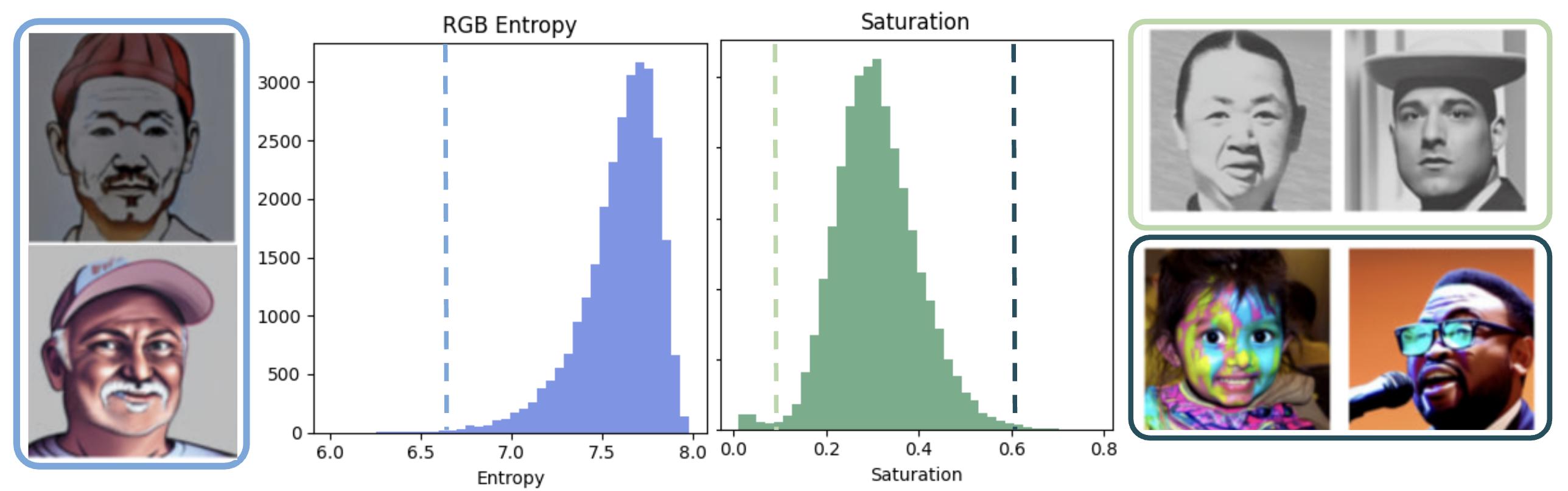}
    \caption{Examples removed by image-statistics filtering. Low entropy corresponds to heavily stylized samples; low saturation identifies grayscale images, while high saturation exposes unnatural colors and lighting artifacts.}
    \label{fig:stats-filter}
\end{figure}

\begin{figure}[bth!]
    \centering
\includegraphics[width=0.95\linewidth]{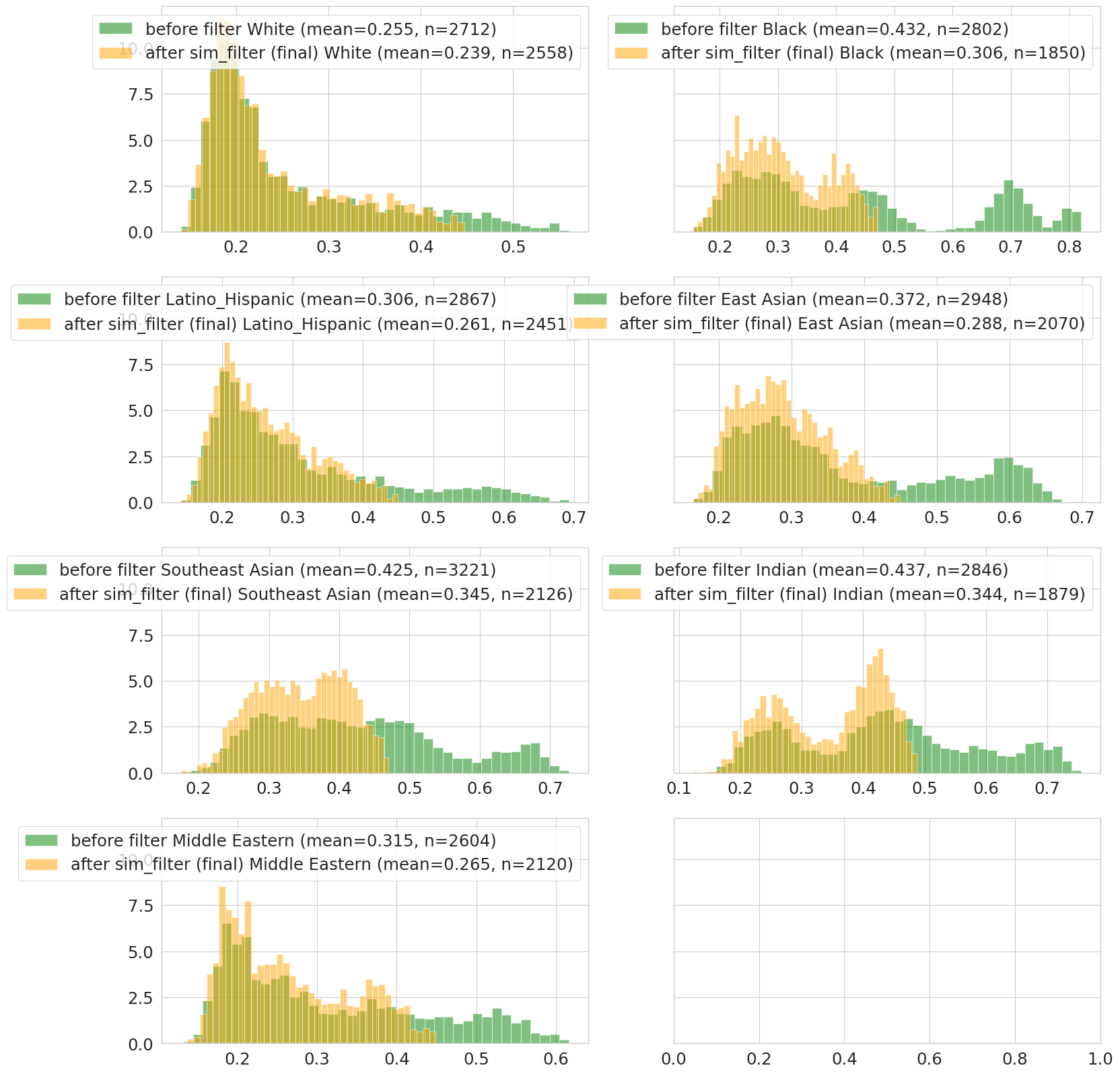}
    \caption{Distribution of similarities within data before and after filtering w.r.t. label of data. Each section correspond to distribution with certain race label attached - effect of filtering eliminates tails present for certain labels}
    \label{fig:simfilter_perlabel}
\end{figure}

In order to measure diversity of embeddings themselves, we analyzed distribution of pairwise distance to closest neighbor within data itself. It can be seen in Figure \ref{fig:simfilter_perlabel} these distributions before and after filtering images based of this metric thresholding - tails of labels that under-perform are dropped, sacrificing ideal uniform distribution of labels for better face diversity. Comparison of diversity metrics to compare before and after are given in Table \ref{tab: filtered sim stats}. The reason for such difference in diversity performance is expected due to inherit bias of underlying training data - even after filtering, certain categories remain unrepresented, which hurts model variability.

\begin{table}[htb]
    \caption{Embedding diversity measured as mean cosine distance to closest neighbor within data. After filtering Stage 2 faces match in quality filtered label data that was used for training and HyperFace baseline}
    \centering
    \begin{tabular}{l l }
        \toprule
        \textbf{Baseline} & \textbf{ID-diversity}  \\
        \midrule
        Real train data & 0.305 \\
        HyperFace & 0.306 \\
        ID-flow samples (before filtering) & 0.355 \\
        \textbf{ID-flow samples (after filtering)} & \textbf{0.295} \\
        \bottomrule
    \end{tabular}
    \label{tab: filtered sim stats}
\end{table}


\subsection{VLM filtering stage}
\label{sec: vlm_filter_stage}

\begin{figure*}[!hbt]
\centering
\begin{promptbox}{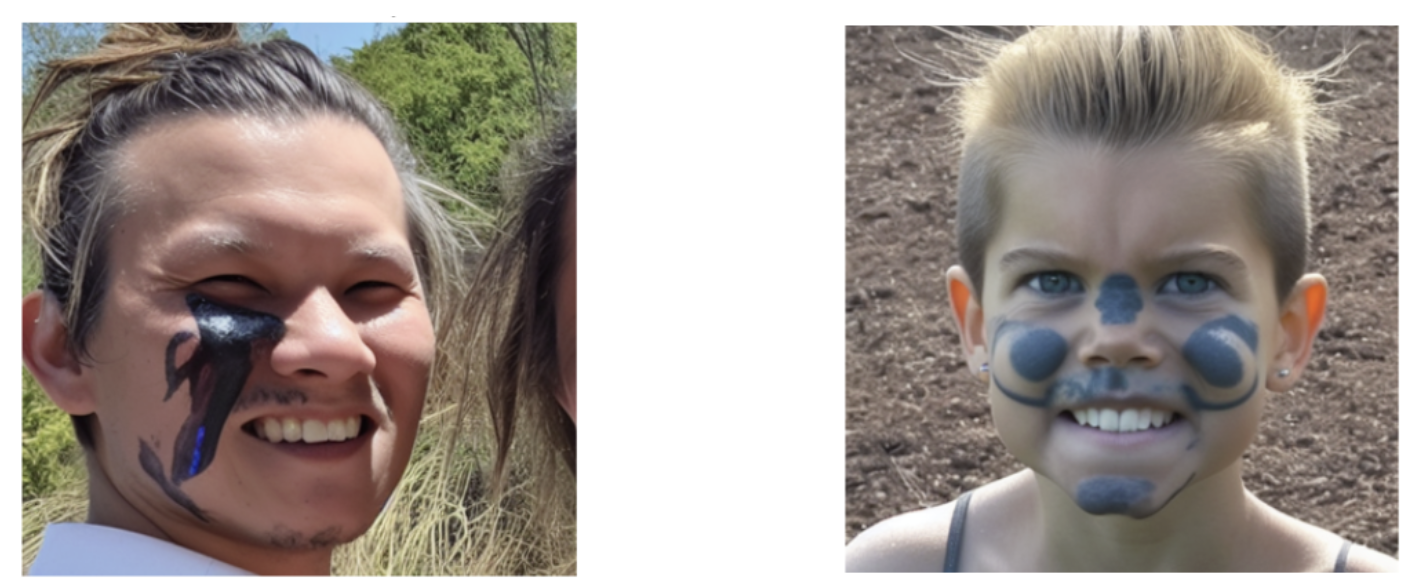}{
   You are given an image of a person. The description of the person is '\{\}' Your task is to visually analyze the image and answer following singular choice questions with options below. Answer all questions binary - 'yes' or 'no'. 
After the answers provide reasoning for answer to Question 2. (about artifacts)
}{
Q1) Does the description of the person roughly match the person in the image? Q2) Does person in the image have any visual artifacts, anomalies or heavy stylization?
}{
1) yes, yes; I think answer to Question 2 is yes because of the black and blue face \textbf{paint} on the person's cheek, which is a form of heavy \textbf{stylization} and a bright/dark color \textbf{spot}. \\
2) no, yes; I think answer to Question 2 is yes because of the face paint applied to the person's face, which includes blue circles around the eyes, and on the cheeks, and on the chin, creating a \textbf{stylized} appearance that suggests a costume or character design.
}
\end{promptbox}
\caption{VLM-filtering output examples of stage 2 - VLM is prompted to visually analyze face in the image, its alignment with labels, detect any artifacts and reason about them}
\label{fig:eval_details}
\end{figure*}

In Figure \ref{fig:eval_details} we provide example of prompt used with Qwen3-VL-30B - specifically with exampels of images where typical artifacts are detected. We highlight keywords in VLM answer that were used to detect positive filtering results and not halucinations related to overall quality of faces generated by Arc2Face.

\subsection{Label matching captions with HQ images}
\label{sec: label_matching_alg}

In Algorithms \ref{alg: class-caption} and \ref{alg: sample-prompts} we describe two-stage strategy to match captions of real images with labels of identities. Firstly, we define a finite set of groups that we classify captions into - these groups are mainly defined by age and gender, e.g. \texttt{adult\_male} and \texttt{child\_female}, and then we collect a final collection of captions for identities by sampling from appropriate group depending on label values of identity. Race labels are intentionally not used for caption matching, so that contextual prompts do not encode race-specific descriptors. This approach has two advantages 1) its algorithmic cost is $O(n +m)$, where $n$ is number of processed captions and $m$ number of processed labels 2) it allows to avoid heavy cases where captions are not suitable for identity, avoiding synthesizing controversial or unnatural content. 
 
\definecolor{commentcolor}{rgb}{0.4,0.4,0.4}
\definecolor{keywordcolor}{rgb}{0.0,0.0,0.7}
\definecolor{stringcolor}{rgb}{0.7,0.0,0.0}

\newcommand{\keyword}[1]{\textcolor{keywordcolor}{\textbf{#1}}}
\newcommand{\codecomment}[1]{\textcolor{commentcolor}{#1}}
\newcommand{\stringliteral}[1]{\textcolor{stringcolor}{#1}}

\begin{algorithm}[ht]
\caption{Process and Classify Captions}
\label{alg: class-caption}
\begin{algorithmic}[1]
\Procedure{ProcessCaptions}{$captions, age\_labels, gender\_labels$}
    \State $categorized\_captions \gets$ \keyword{empty dictionary}
    
    \ForAll{$(caption, age, gender)$ in $(captions, age\_labels, gender\_labels)$}
        \State \codecomment{Determine target category for caption (by key-words and LLM processing):}
        \State \hspace{\algorithmicindent} \codecomment{e.g. \stringliteral{baby}, \stringliteral{child} (male/female), or \stringliteral{adult} (male/female)}
        \State \codecomment{Sample random age-appropriate person-descriptor (e.g., \stringliteral{child} vs \stringliteral{adult})}
        \State \codecomment{Replace generic person descriptors with age-appropriate terms (not just man/woman)}
        \State \codecomment{Store processed caption under its target category}
    \EndFor
    
    \State \keyword{return} $categorized\_captions$
\EndProcedure
\end{algorithmic}
\end{algorithm}

\begin{algorithm}[ht]
\caption{Sample prompts for identity labels}
\label{alg: sample-prompts}
\begin{algorithmic}[1]
\Procedure{SamplePrompts}{$age\_labels, gender\_labels, categorized\_captions$}
    \State $prompts \gets$ \keyword{empty list}
    
    \ForAll{$(age, gender)$ in $(age\_labels, gender\_labels)$}
        \State \codecomment{Match (age, gender) pair to suitable category (child, adult, etc.)}
        \State \codecomment{Fetch next available caption from that category}
        \State \codecomment{Append caption to prompts list}
    \EndFor
    
    \State \keyword{return} $prompts$
\EndProcedure
\end{algorithmic}
\end{algorithm}

\subsection{In-context model for human generation analysis}
\label{sec: incontext_ablation}

\begin{figure}[bth!]
    \centering
    \includegraphics[width=0.9\textwidth]{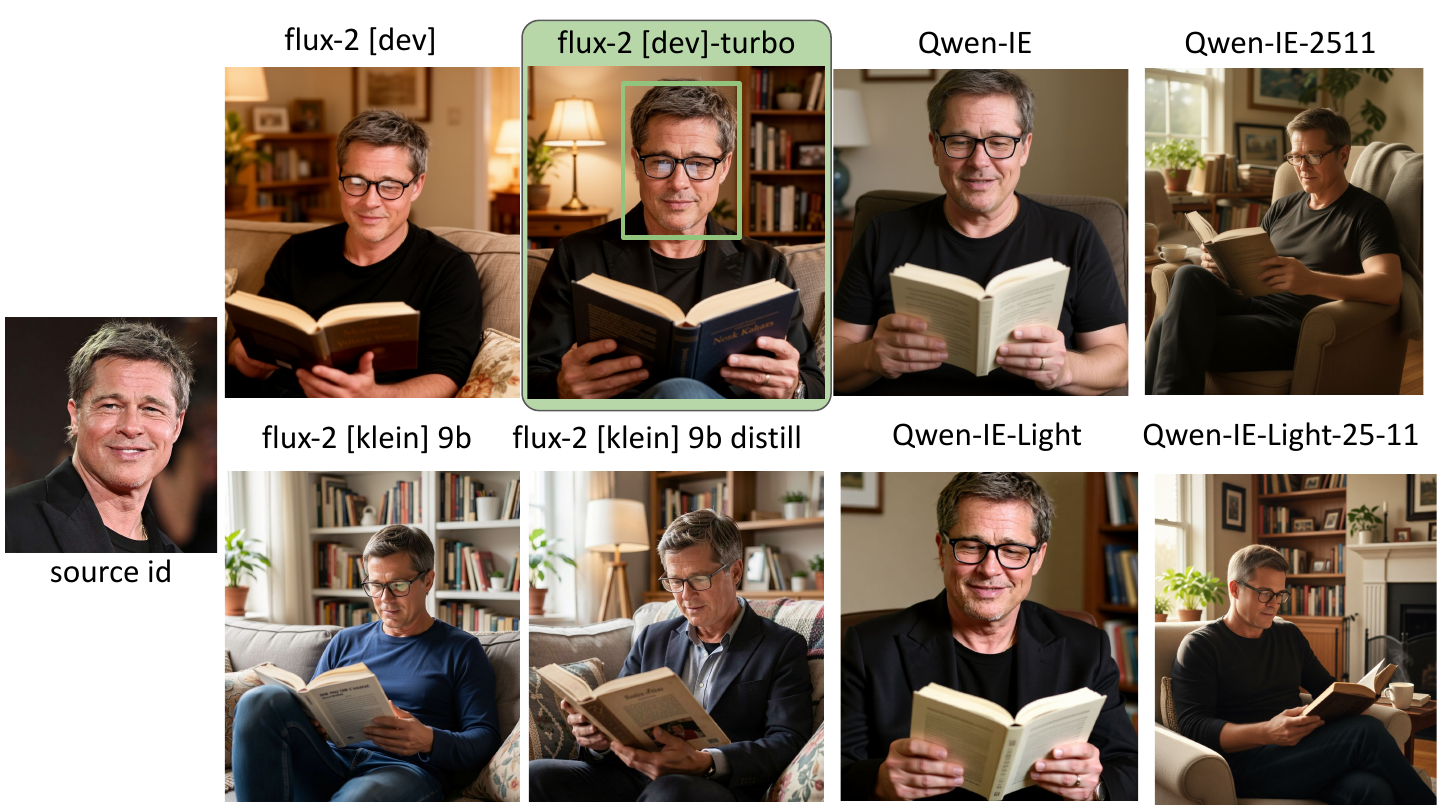}
    \caption{Comparison of various models and checkpoints of Flux-2 and Qwen-IE for contextualization stage. It can be seen that only \textbf{Flux-2 [dev]- turbo} achieves both preservation and fidelity. Images generated with same prompt "Realistic image of man from image 1, in glasses reading book at home"}
    \label{fig:comparison}
\end{figure}

In order to choose optimal latest model for \textbf{Stage 3} we manually analyzed various modern pipelines that are suited for generation while preserving identity face - including Qwen Image and latest Flux 2.0 models. Despite these models being understudied for this application, some significant challenges arise for these types models - especially character consistency with respect to prompt. Figure \ref{fig:comparison} shows that for most pipelines realistic preservation of face remains open problem and is largely dependent on prompt.

\subsection{ID-preserving generation details and results}
\label{sec: finetune_details}

\begin{figure}[bth!]
    \centering
    \includegraphics[width=0.95\textwidth]{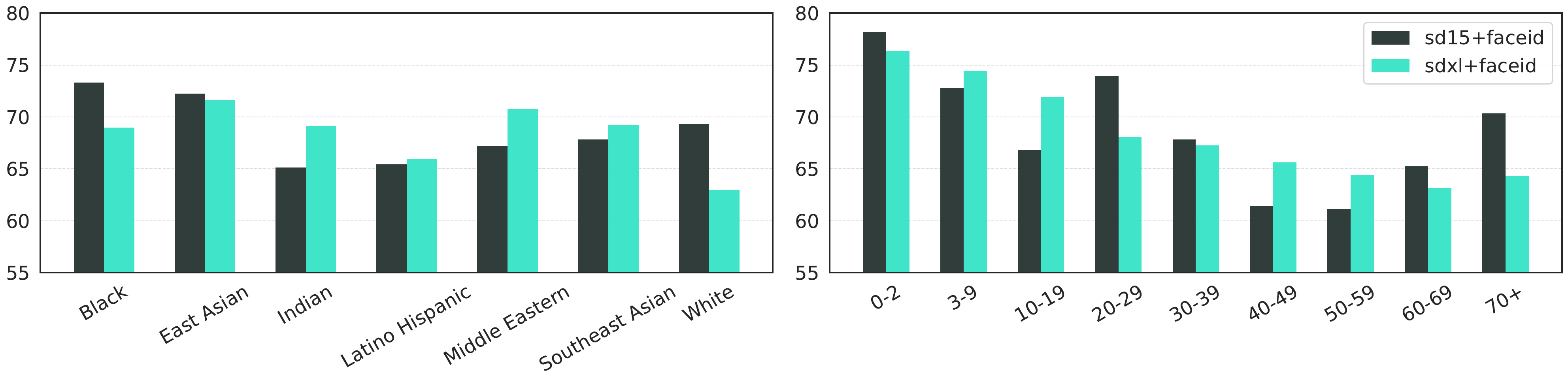}
    \caption{Face similarity metric unfairness for different age categories in identity preserving generation - measured preservation across labels is far from balanced}
    \label{fig:facesim-unfairness}
\end{figure}

In Figure \ref{fig:facesim-unfairness} we highlight performance imbalance of some conventional pipelines, where for different categories face preservation may significantly deviate.  

\paragraph{SD-1.5+FaceID} For finetuning IpAdapter-FaceID we additionally process collected dataset. Since this adapter uses global diffusion loss, we filter out images with more then one human in the image. To enable shortcut finetuning for selected images we additionally extract openpose control signals. We do not train image projector layers and only finetune blocks of IpAdapter added inside StableDiffusion Unet. We use \texttt{Adam-W} optimizer with \texttt{lr}=$3e-5$ and do not apply warmup, \texttt{bs}=8. In total we train model for 2500 iterations. 

\begin{figure}[bth!]
    \centering
    \includegraphics[width=0.9\textwidth]{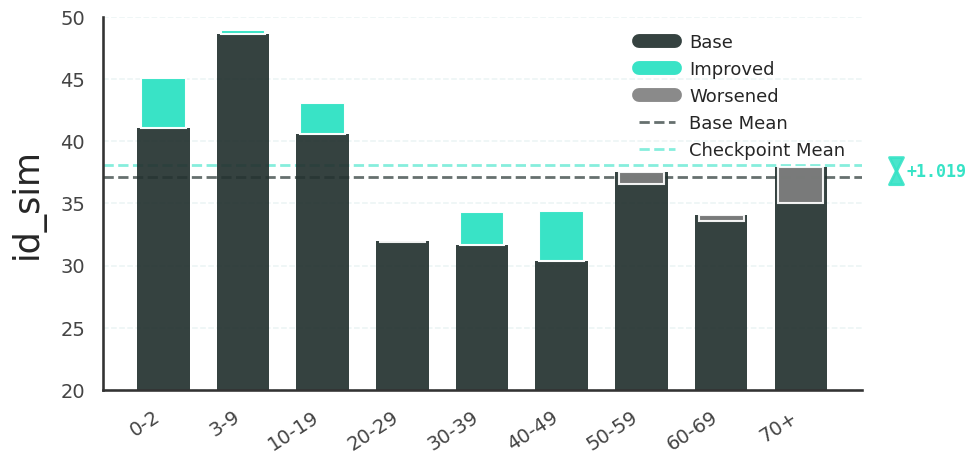}
    \caption{Finetuning results for Photomaker - \textbf{FairBench} benchmark shows average improvement of quality, while also balancing performance between categories}
    \label{fig:photomaker-tune}
\end{figure}

\paragraph{SDXL+Photomaker} In order to finetune Photomaker we do not rely on shortcut finetuning, but instead follow training recipe described in original paper. We modify collected captions by ensuring trigger-word is added single time next to person descriptor.  We finetune LoRA modules attached to linear layers in attention blocks with \texttt{rank}=16  with \texttt{Adam-W} optimizer, learning rate set to $1e-5$, use linear warm up for approximately 2k iterations and train in total for 20000 iterations with \texttt{bs}=4. In Figure \ref{fig:photomaker-tune} we provide result of finetuning with our dataset - it has a narrower effect of improvement then result for FaceID, which can be attributed to complexity of finetuning task with this model.

\subsection{More samples from SynthFaces dataset}
\label{sec: synthface samples}

\begin{figure}[!hb]
    \centering
    \includegraphics[width=0.95\textwidth]{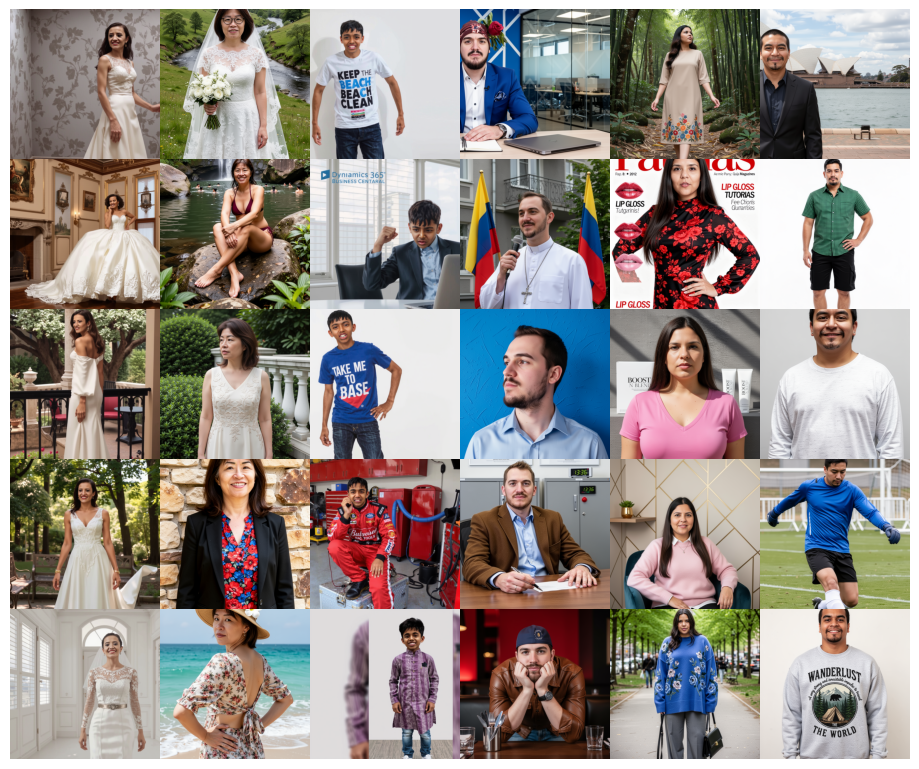}
    \caption{More samples from SynthFaces dataset, including non-square resolution images cropped around faces. Each column corresponds to images of the same synthetic identity.}
    \label{fig:id-grouped}
\end{figure}

In Fig. \ref{fig:id-grouped} we show more HQ samples from synthfaces dataset, grouped by same identity.

\subsection{Fairness Evaluation: Supplementary Tables}
\label{sec:fairness_appendix}

\paragraph{Demographic distribution in standard FR benchmarks.}

Tables~\ref{tab:benchmark_race_dist}. quantifies the racial skew in standard FR benchmarks, annotated with FairFace~\cite{karkkainen2021fairface}. The heavy White-male dominance motivates the use of SER over accuracy-based fairness metrics. Other tables showcase global accuracy and SER metrics on FR benchmarks.

\begin{sidewaystable}[!htbp]
    \centering
    \caption{Race distribution (\% of pairs) in standard FR benchmarks annotated with FairFace.}
    \label{tab:benchmark_race_dist}
    \resizebox{0.75\textwidth}{!}{
    \begin{tabular}{lccccccc}
        \toprule
        \textbf{Benchmark} & White & M.~Eastern & Black & E.~Asian & Latino & Indian & SE~Asian \\
        \midrule
        LFW      & \textbf{65.9} & 15.9 & 6.7  & 5.4 & 3.7 & 1.0 & 1.5 \\
        CFP-FP   & \textbf{67.4} & 9.0  & 11.6 & 1.7 & 5.7 & 3.9 & 0.8 \\
        AgeDB-30 & \textbf{88.0} & 7.8  & 0.9  & 0.4 & 2.3 & 0.6 & $<$0.1 \\
        CALFW    & \textbf{68.7} & 13.8 & 7.1  & 5.2 & 3.5 & 0.9 & 0.7 \\
        CPLFW    & \textbf{62.7} & 19.6 & 7.7  & 5.6 & 2.8 & 0.9 & 0.8 \\
        \bottomrule
    \end{tabular}
    }
\end{sidewaystable}

\begin{table}[htb]
    \centering
    \caption{Global accuracy and AUC on standard FR benchmarks ($\uparrow$\,better). SynthFaces-500K's lower accuracy reflects its earlier training stage.}
    \label{tab:global_acc_fr}
    \resizebox{\textwidth}{!}{
    \begin{tabular}{lcccccccccccc}
        \toprule
        & \multicolumn{2}{c}{LFW} & \multicolumn{2}{c}{CFP-FP} & \multicolumn{2}{c}{AgeDB-30} & \multicolumn{2}{c}{CALFW} & \multicolumn{2}{c}{CPLFW} & \multicolumn{2}{c}{\textbf{Mean}} \\
        \cmidrule(lr){2-3}\cmidrule(lr){4-5}\cmidrule(lr){6-7}\cmidrule(lr){8-9}\cmidrule(lr){10-11}\cmidrule(lr){12-13}
        \textbf{Model} & Acc & AUC & Acc & AUC & Acc & AUC & Acc & AUC & Acc & AUC & \textbf{Acc} & \textbf{AUC} \\
        \midrule
        SynthFaces-500K        & 0.913 & 0.971 & 0.727 & 0.803 & 0.713 & 0.791 & 0.791 & 0.867 & 0.679 & 0.741 & 0.764 & 0.835 \\
        Vec2Face  & 0.993 & 0.999 & 0.913 & 0.969 & 0.945 & 0.984 & 0.946 & 0.973 & 0.877 & 0.927 & 0.935 & 0.970 \\
        HyperFace   & 0.983 & 0.998 & 0.880 & 0.945 & 0.855 & 0.931 & 0.879 & 0.944 & 0.825 & 0.891 & 0.884 & 0.942 \\
        \bottomrule
    \end{tabular}
    }
\end{table}

\begin{table}[htb]
    \centering
    \caption{Maximum racial SER at FPR\,=\,1\% on standard FR benchmarks ($\downarrow$\,fairer). $\star$\,=\,SynthFaces-500K is fairest.}
    \label{tab:ser_fpr1}
    \resizebox{\textwidth}{!}{
    \begin{tabular}{lcccccr}
        \toprule
        \textbf{Model} & LFW & CFP-FP & AgeDB-30 & CALFW & CPLFW & \textbf{Mean} \\
        \midrule
        SynthFaces-500K (ours) & \textbf{1.482} & \textbf{1.133} & \textbf{1.117} & \textbf{1.087} & \textbf{1.040} & \textbf{1.172} \\
        Vec2Face-HSFace300K   & 1.536 & 1.347 & 1.670 & 1.196 & 1.344 & 1.419 \\
        HyperFace-10K-LDM     & 2.061 & 1.155 & 1.686 & 1.032 & 1.137 & 1.414 \\
        \bottomrule
    \end{tabular}
    }
\end{table}

\end{document}